\documentclass{article}

\usepackage{iclr2027_conference,times}
\usepackage{url}

\usepackage{latexsym}

\usepackage[T1]{fontenc}

\usepackage[utf8]{inputenc}

\usepackage{microtype}

\usepackage{inconsolata}

\usepackage{graphicx}
\usepackage{wrapfig}

\usepackage{enumitem}
\usepackage{tabularx}
\usepackage{multirow}
\usepackage{booktabs}
\usepackage{tikz}
\usepackage{hyperref}
\usepackage[table]{xcolor}
\usepackage{arydshln}
\usepackage{amsmath}
\usepackage{amsfonts}
\usepackage{amssymb} 
\usepackage{CJKutf8}
\usepackage{algorithm}
\usepackage{algpseudocode}

\iclrfinalcopy

\title{\Large From Anomalies to Failures: Constructing\\
Causal Error Graphs for Agentic Trace Diagnosis}

\author{
Shu-Xun Yang$^{1,2}$\thanks{Work done during an internship at Zhipu AI.} \quad
Yidong Wang$^{2}$ \quad
Zhuoer Feng$^{2,3}$ \quad
Bosi Wen$^{2,3}$ \quad
Jiayi Gui$^{2}$ \\
\textbf{Dayong Yang}$^{2}$ \quad
\textbf{Wenbo Yu}$^{2}$ \quad
\textbf{Haoke Zhang}$^{2}$ \quad
\textbf{Jie Tang}$^{3}$ \quad
\textbf{Cunxiang Wang}$^{2,3}$\thanks{The corresponding author} \\
$^{1}$Beijing Institute of Technology, Beijing, China \\
$^{2}$Zhipu AI, Beijing, China \quad
$^{3}$Tsinghua University, Beijing, China \\
\texttt{sheryl.xun@bit.edu.cn; wangcunxiang303@gmail.com}
}

\vspace{-5mm}

\begin{document}
\pagestyle{fancy}
\fancyhead{}
\renewcommand{\headrulewidth}{0pt}
\maketitle
\begin{abstract}

LLM-driven agents are increasingly deployed in complex applications, where long agentic traces make failures difficult to diagnose. Existing trace diagnosis methods often conflate anomalies, errors, and failures, making diagnostic targets ambiguous; they also lack structured modeling of how causally relevant errors propagate and amplify into final task failures, resulting in unreliable failure attribution. To address these problems, we propose \textbf{CEG-Agent}, a tool-augmented agentic framework for causal diagnosis of agentic traces. Specifically, CEG-Agent introduces an explicit taxonomy of anomalies, errors, and failures, and constructs \textbf{Causal Error Graphs} (CEGs), a unified typed representation that links execution events, diagnostic nodes, and failure outcomes through causal relations. 
To evaluate causal trace diagnosis, we further construct \textbf{CEG-Bench}, a fully agent-annotated benchmark with high-confidence, consensus-derived CEG annotations obtained through an \textbf{Adversarial Agentic Adjudication Protocol} (AAAP). We validate the resulting annotations against an expert-curated human gold set, which shows close agreement with the automatic annotations. Experiments on CEG-Bench demonstrate that CEG-Agent achieves state-of-the-art performance under both semantically relaxed and structurally exact evaluation criteria.
Our code is publicly available \footnote{https://github.com/SHU-XUN/CEG-Agent}.


\end{abstract}

\section{Introduction}

\begin{wrapfigure}{r}{0.46\textwidth}
  \vspace{-1cm}
  \centering
  \vspace{-\intextsep}
  \includegraphics[width=0.38\textwidth]{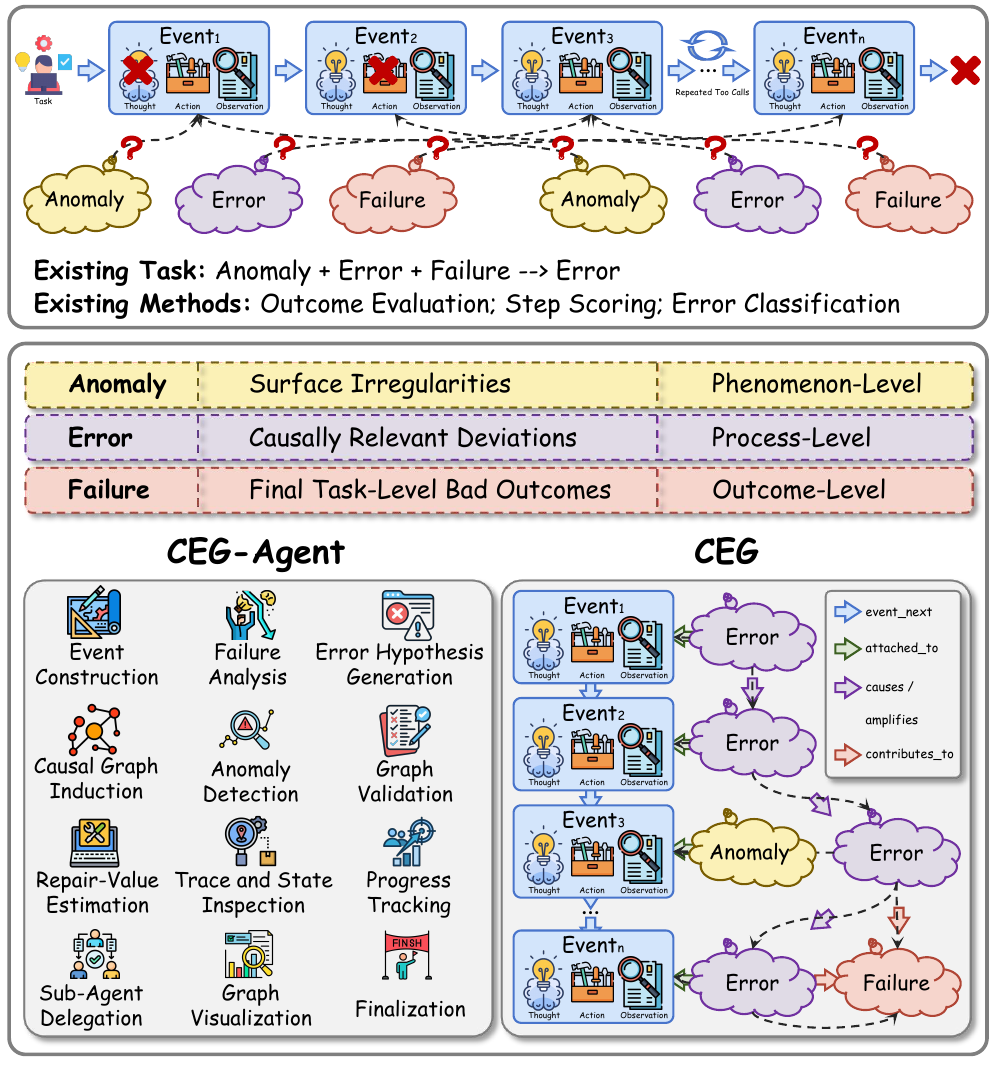}
  \vspace{-0.2cm}
  \caption{Comparison between prior paradigm and our method for agentic trace diagnosis.}
  \label{fig1}
  \vspace{-0.4cm}
\end{wrapfigure}

In recent years, LLM-driven agents have been increasingly deployed in complex applications such as deep research \citep{browsecomp-deepresearch}, function calling \citep{tau2-functioncalling}, and code generation \citep{swebench-coding}, where executions often unfold into long agentic traces with multiple intermediate steps and interactions. While effective, these long traces are prone to diverse failures \citep{tracesir}. Such failures often arise from causally relevant process-level errors that gradually propagate and amplify across the trace, rather than from surface anomalies or isolated local errors. This highlights the need for automatically diagnosing failures in agentic traces and their underlying causal errors to support agent debugging and improvement.

However, existing methods for agentic trace diagnosis still suffer from two fundamental limitations. First, they often collapse heterogeneous diagnostic signals in agentic traces under the generic label of \textbf{error}. As shown in Figure \ref{fig1}, a failed trace may exhibit (1) surface irregularities, such as \textit{repeated tool calls} or \textit{unnecessarily verbose reasoning}; (2) causally relevant deviations, such as \textit{missing constraints} or \textit{improper tool use}; and (3) final task-level bad outcomes, such as \textit{incorrect answers} or \textit{incomplete tasks}. These heterogeneous signals naturally call for a distinction among \textbf{anomalies}, \textbf{errors}, and \textbf{failures}, which correspond to phenomenon-level irregularities, process-level deviations, and outcome-level breakdowns, respectively. Conflating these notions as generic errors obscures their distinct diagnostic roles and optimization targets, making it difficult to formulate trace diagnosis as a well-defined problem and to provide consistent and effective supervision signals.
Second, existing approaches rarely capture the causal structure of errors within a trace. They mostly rely on flat error classification \citep{codetracer} or step scoring \citep{stepmath}, lacking a structured account of how errors propagate, amplify, and eventually lead to failures. As a result, diagnosis often struggles to identify the key causes that truly drive the failure and the propagation paths where they take effect. 
This in turn weakens failure attribution and hinders debugging and optimization, since developers cannot confidently determine where to intervene first or which deviations are most worth correcting.

To address these limitations, we introduce \textbf{CEG-Agent}, a tool-augmented agentic framework for causal diagnosis of agentic traces. CEG-Agent starts from an explicit diagnostic taxonomy that separates \textbf{anomalies}, \textbf{errors}, and \textbf{failures}: anomalies capture phenomenon-level irregularities, errors capture causally relevant process-level deviations, and failures capture outcome-level task breakdowns. Based on this taxonomy, we define the \textbf{Causal Error Graph} (CEG) as a unified typed representation over execution events, anomalies, errors, and failures, with causal edges that specify where errors occur, how they propagate or amplify one another, and how they contribute to task-level failures. To automatically construct such graphs, CEG-Agent composes a set of core diagnosis tools for event construction, failure analysis, error hypothesis generation, causal graph induction, anomaly detection, graph validation, and repair-value estimation, together with auxiliary tools for trace and state inspection, progress tracking, sub-agent delegation, graph visualization, and finalization. The resulting system produces visualizable causal error graphs with repair-value estimates, helping developers understand failure formation and prioritize interventions for agent improvement.

To evaluate causal trace diagnosis, we construct \textbf{CEG-Bench}, a fully agent-annotated benchmark with high-confidence, consensus-derived CEG annotations. Following TraceSIR \citep{tracesir}, we build a candidate pool of 150 failed GLM-4.6 traces from BrowseComp \citep{browsecomp-deepresearch}, Tau2Bench \citep{tau2-functioncalling}, and SWE-bench \citep{swebench-coding}, with 50 cases from each benchmark. Given the complexity of agentic traces and the difficulty of manual causal annotation, we design an \textbf{Adversarial Agentic Adjudication Protocol} (AAAP) for CEG annotation. Specifically, Claude Code powered by Claude-Opus-4.7 first drafts an initial CEG for each candidate trace. The draft is then revised using diagnostic feedback from the open-source TraceSIR analyzer powered by Claude-Opus-4.6, which serves as an independent error reviewer. Next, two heterogeneous agentic annotators, Claude Code with Claude-Opus-4.7 and Codex with GPT-5.5, iteratively cross-examine and refine each graph until both reach consensus and raise no remaining objections to the CEG. We finally retain the 100 cases whose CEGs are unanimously accepted by these two strong agentic frameworks powered by frontier LLMs, yielding a final set of 100 consensus traces.

We validate these automatically constructed annotations against \textbf{CEG-Bench-Gold}, an expert-curated human reference set over the same 100 traces. The AAAP annotations show strong agreement with human judgments, achieving an overall $\mathrm{CEG\text{-}Sim}$ of $71.5$ and high agreement on core diagnostic dimensions. This agreement provides empirical support for AAAP as a scalable and reliable annotation protocol for causal trace diagnosis.
Furthermore, experiments on \textbf{CEG-Bench} show that \textbf{CEG-Agent} outperforms strong agentic baselines, including Claude Code and Codex, with average gains of $+9.0$ points under \textbf{semantically relaxed} CEG evaluation and $+4.9$ points under \textbf{structurally exact} CEG evaluation. Fine-grained analyses show that these gains are concentrated on the core structural capabilities required for causal diagnosis, such as distinguishing anomalies from causally relevant errors and constructing error propagation chains. Cross-backbone results further show that CEG-Agent exhibits different diagnostic profiles when driven by different LLMs, yet consistently provides explicit, structured, and causally grounded diagnoses for debugging and improvement.

\begin{figure*}[t]
  \centering
  \includegraphics[width=\linewidth]{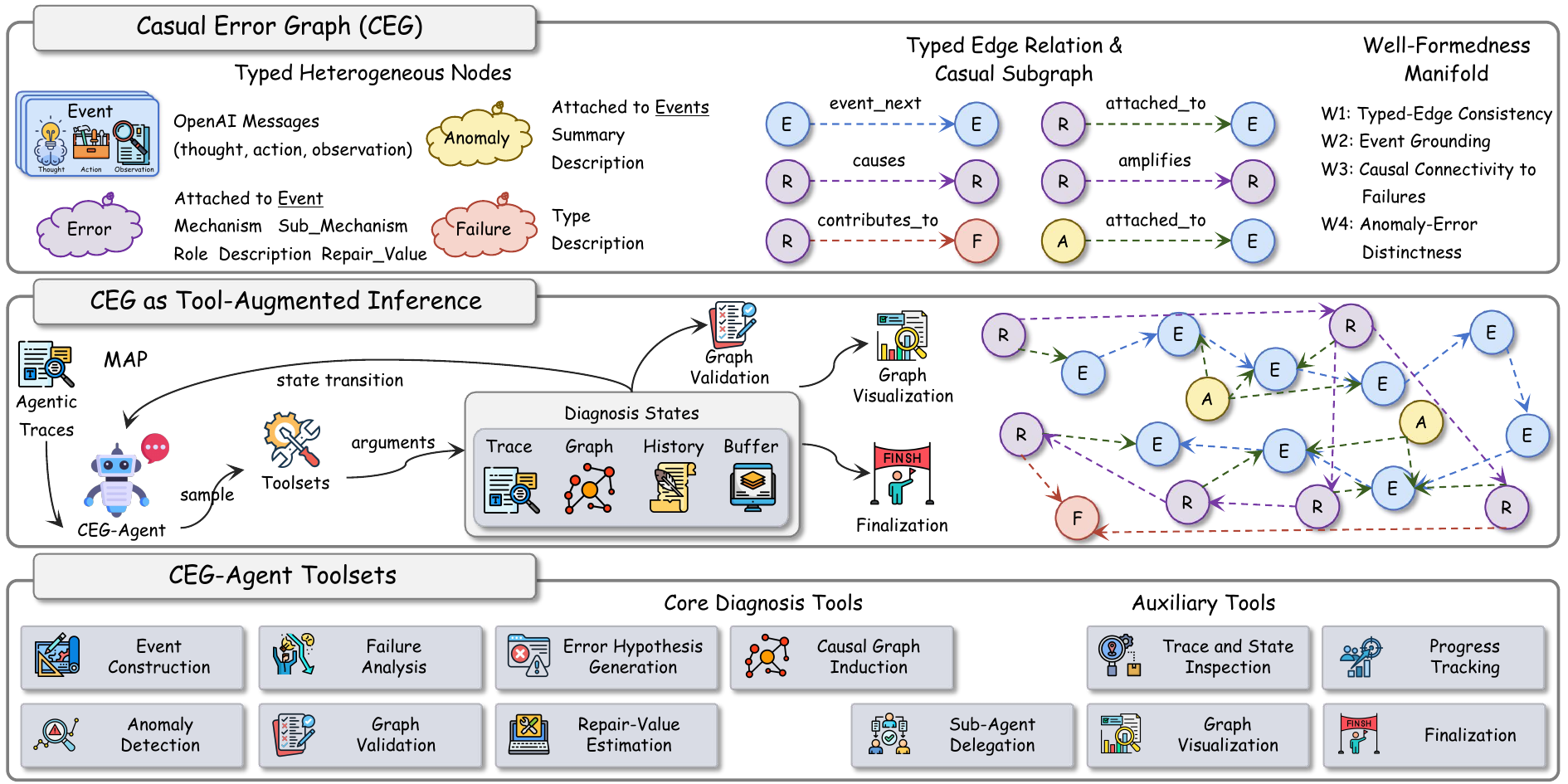}
  \vspace{-6mm}
  \caption{The overall design of CEG-Agent.}
  \label{fig2}
  \vspace{-4mm}
\end{figure*}

\section{Related Work}

LLM-driven agents are increasingly evaluated on long-horizon tasks involving web browsing, deep research, function calling, tool use, and software engineering. Benchmarks such as BrowseComp \citep{browsecomp-deepresearch}, Tau2Bench \citep{tau2-functioncalling}, SWE-bench \citep{swebench-coding}, and broader agent benchmarks \citep{agentbench,webarena,mind2web} have advanced evaluation of task completion and tool-use capabilities, while process-oriented evaluations and LLM-based judges further assess intermediate reasoning, actions, and step-level quality \citep{llmjudge,processreward}. Recent trace-level diagnosis methods, especially TraceSIR \citep{tracesir}, begin to analyze agentic execution traces through structured reports, trajectory feedback, or error detection, and are related to work on hallucination detection, reasoning errors, tool-use failures, and software fault diagnosis \citep{toolformer,codetracer,stepmath}. However, these methods often focus on final outcomes, local step quality, or flat error categories, and provide limited support for distinguishing surface anomalies, causally relevant process-level errors, and outcome-level failures, or for explaining how errors propagate and amplify into failures. Our work is also connected to causal reasoning, graph-based diagnosis, provenance modeling, and root cause analysis, where causal graphs and dependency paths are used to explain failures \citep{csnake,cascading,causalexplainability,failuremodes}. CEG adapts these ideas to LLM-driven agents by grounding process-level errors in execution events and linking them to failures through typed causal relations. Finally, CEG-Agent builds on tool-augmented and critic-based LLM frameworks, including ReAct-style tool use, self-refinement, reflection, and multi-agent collaboration \citep{react,selfrefine,reflexion}, but specializes them for structured causal trace diagnosis, producing visualizable CEGs with repair-value estimates for debugging and agent improvement.


\section{Method}
\label{sec:method}

We present \textbf{CEG-Agent}, a tool-augmented agentic framework that formulates agentic trace diagnosis as constrained inference over typed heterogeneous causal graphs, as shown in Figure \ref{fig2}. Specifically, we first formalize causal trace diagnosis as a constrained \emph{maximum a posteriori} (MAP) problem, where the target is a schema-valid graph explaining task-level failures and their underlying causally relevant errors from an agentic trace (\S\ref{sec:method-problem}). We then define the \textbf{Causal Error Graph} (CEG) as a heterogeneous multi-relational graph that separates anomalies, errors, and failures, grounds process-level deviations in execution events, and encodes error propagation and failure attribution through typed edges (\S\ref{sec:method-ceg}). Finally, we instantiate CEG construction as a tool-augmented inference procedure, where a diagnostic agent incrementally constructs, validates, and revises the graph through schema-respecting state transitions (\S\ref{sec:method-inference}). Semantic refinement, graph validation, and counterfactual repair-value estimation are also exposed as registered tools in this process. This section focuses on the core modeling and inference principles, while implementation details are given in Appendix~\ref{sec:impl}.

\subsection{Problem Formulation}
\label{sec:method-problem}

An \emph{agentic trace} is a pair $T = (\tau, \mathcal{M})$ where $\tau$ is the user task specification and $\mathcal{M} = (m_1, \ldots, m_N)$ is a sequence of OpenAI-style messages spanning assistant reasoning, tool calls, and observations. Trace diagnosis is the problem of inferring a \emph{causal error graph} $G \in \mathcal{G}^\star(T)$, where $\mathcal{G}^\star(T)$ denotes the schema-valid set of graphs grounded in $T$. We model this problem as constrained MAP estimation:
\begin{equation}
G^\star \;=\; \arg\max_{G\,\in\,\mathcal{G}^\star(T)} \;\log p_\theta(G \mid T),
\label{eq:map}
\end{equation}
where $p_\theta$ is implicitly defined by a tool-augmented agentic policy (\S\ref{sec:method-inference}). The constraint $G\in\mathcal{G}^\star(T)$ enforces three diagnostic requirements that distinguish CEG inference from flat error classification: (R1) typed separation of phenomenon-level anomalies, process-level errors, and outcome-level failures; (R2) grounding of each diagnostic node in a concrete execution event; (R3) explicit causal structure linking errors to the failure they explain.

\subsection{Causal Error Graph}
\label{sec:method-ceg}

We formalize a CEG as a typed heterogeneous multi-relational graph $G \;=\; \bigl(V,\, E,\, \tau_V,\, \tau_E,\, \mathbf{x}\bigr)$, with a node-type map $\tau_V:V\!\rightarrow\!\mathcal{T}_V$, an edge-type map $\tau_E:E\!\rightarrow\!\mathcal{L}$, and an attribute map $\mathbf{x}$ that assigns each node a type-specific attribute tuple.

\paragraph{Heterogeneous node types.}
The node-type alphabet is $\mathcal{T}_V = \{\mathrm{E},\mathrm{A},\mathrm{R},\mathrm{F}\}$, inducing disjoint heterogeneous subsets $V \;=\; V_E \,\dot{\cup}\, V_A \,\dot{\cup}\, V_R \,\dot{\cup}\, V_F$, where $V_E$ contains \emph{event} nodes, $V_A$ contains \emph{anomaly} nodes, $V_R$ contains \emph{error} nodes, and $V_F$ contains \emph{failure} nodes. Attribute tuples are $\mathbf{x}(e) = (\textit{thought},\textit{action},\textit{observation})$ for $e\in V_E$; $\mathbf{x}(r) = (\varepsilon(r), \mu(r), \nu(r), \rho(r), d(r))$ for $r\in V_R$ with grounding event $\varepsilon(r)\in V_E$, coarse mechanism $\mu(r)\in\mathrm{Mech}$, fine sub-mechanism $\nu(r)$, structural role $\rho(r)\in\{\textsc{root},\textsc{propagated},\textsc{amplification}\}$, and a trace-grounded description $d(r)$; and $\mathbf{x}(f) = (\theta(f), d(f))$ for $f\in V_F$ with failure type $\theta(f)$. The mechanism ($8$ values), role ($3$ values), and failure-type ($6$ values) vocabularies are \emph{grounded} in large-scale human error analyses, and \emph{extensible}, with an explicit \emph{other} bucket as a completeness escape hatch; empirically all mechanisms and all failure types are populated in CEG-Bench, with \emph{other} absorbing only $3$ residual failures (\ref{app:bench-vocab}).
$V_A$ may be grounded in one or more events, denoted by $\mathbf{x}(a) = (\boldsymbol{\varepsilon}(a)\subseteq V_E, \nu(a), d(a))$.

\paragraph{Typed edge relation.}
The edge-type alphabet is 
$\mathcal{L} = \{$\textsc{event-next}, \textsc{attached-to}, \textsc{causes}, \textsc{amplifies}, \textsc{contributes-to}$\}$.
To enforce heterogeneity, every edge respects a legal endpoint relation $\Omega \subset \mathcal{T}_V \times \mathcal{L} \times \mathcal{T}_V$:
\begin{equation}
\forall (u,\ell,v)\in E:\;\bigl(\tau_V(u),\,\ell,\,\tau_V(v)\bigr)\in\Omega,
\label{eq:typed-edge}
\end{equation}
where $\Omega$ is the valid type tuples:
$(\mathrm{E}, \textsc{event-next}, \mathrm{E})$, 
$(\mathrm{R}, \textsc{attached-to}, \mathrm{E})$, 
$(\mathrm{R}, \textsc{causes}, \mathrm{R})$, 
$(\mathrm{R}, \textsc{amplifies}, \mathrm{R})$, and 
$(\mathrm{R}, \textsc{contributes-to}, \mathrm{F})$.
Anomalies in $V_A$ may also be grounded in one or more events via \textsc{attached-to} edges. Since they are surfaced for diagnostic completeness but excluded from causal attribution, we do not emphasize these edges in the following discussion.

\paragraph{Causal subgraph and reachability.}
Let $\mathcal{L}_C = \{$\textsc{causes}, \textsc{amplifies}, \textsc{contributes-to}$\}$ denote the causal sub-alphabet, and let $G_C = (V_R \cup V_F, E_C)$, where $E_C = \{(u,\ell,v) \in E \mid \ell \in \mathcal{L}_C\}$, be the \emph{causal subgraph} obtained by projecting away temporal and grounding edges. For an error $r \in V_R$, we define the failure reachability set under $G_C$ as $\mathrm{Reach}_G(r) = \{f \in V_F \mid r \xrightarrow{\mathcal{L}_C^{\ast}} f\}$, where $\xrightarrow{\mathcal{L}_C^{\ast}}$ denotes any directed path using only causal edge types.

\paragraph{Well-formedness manifold.}
The manifold $\mathcal{G}^\star(T)$, which acts as the feasible region for the diagnostic optimization problem in Eq.~\eqref{eq:map}, is the set of graphs satisfying four predicates. $\mathrm{W}_1$: typed-edge consistency, i.e., Eq.~\eqref{eq:typed-edge}; $\mathrm{W}_2$: event grounding, requiring $\forall r\in V_R,\exists e\in V_E:(r,\textsc{attached-to},e)\in E$; $\mathrm{W}_3$: causal connectivity to failures, ensuring $\forall r\in V_R,\mathrm{Reach}_G(r)\neq\varnothing$; and $\mathrm{W}_4$: anomaly--error distinctness, $\forall a\in V_A,\forall r\in V_R: a \not\equiv r$, forbidding textually equivalent diagnostic content.


\subsection{CEG as Tool-Augmented Inference}
\label{sec:method-inference}


We approximate the MAP objective in Eq.~\eqref{eq:map} by a constrained tool-augmented inference procedure. A top-level LLM policy maintains a diagnosis state and incrementally constructs $G$ by invoking specialized tools, each of which writes to one component of the state under the schema constraints of $\mathcal{G}^\star(T)$. Let $S_t=(T,\,G_t,\,H_t,\,\mathcal{Q}_t)$ denote the diagnosis state at iteration $t$, comprising the trace $T$, the current partial graph $G_t$, the conversation history $H_t$, and an auxiliary work-tracking buffer $\mathcal{Q}_t$ containing todo list and sub-agent reports. At every step, the agent samples a tool $u_t$ from the registered tool set $\mathcal{U}$ with its execution arguments $\mathbf{a}_t$, and induces a state transition. 
A distinguished terminal action $u_T = \texttt{finalize}$ halts the trajectory; otherwise the loop runs for at most $K$ iterations. 
\begin{equation}
\bigl(u_t,\, \mathbf{a}_t\bigr) \sim \pi_\theta(\,\cdot\mid S_t,\,\mathcal{U}),\;\;
S_{t+1} = \Phi_{u_t}\!\bigl(S_t,\,\mathbf{a}_t\bigr).
\label{eq:policy-transition}
\end{equation}

Specifically, the tool set $\mathcal{U}$ of CEG-Agent is designed to mirror the diagnostic structure presented in \S\ref{sec:method-ceg}. It consists of seven \emph{core diagnosis tools} that construct, validate, and value the components of the CEG, and five \emph{auxiliary tools} that support the diagnostic process without directly writing graph components.
We outline high-level role of each tool below, deferring implementation details to \S\ref{sec:impl}. 

\begin{figure*}[t]
  \centering
  \includegraphics[width=\linewidth]{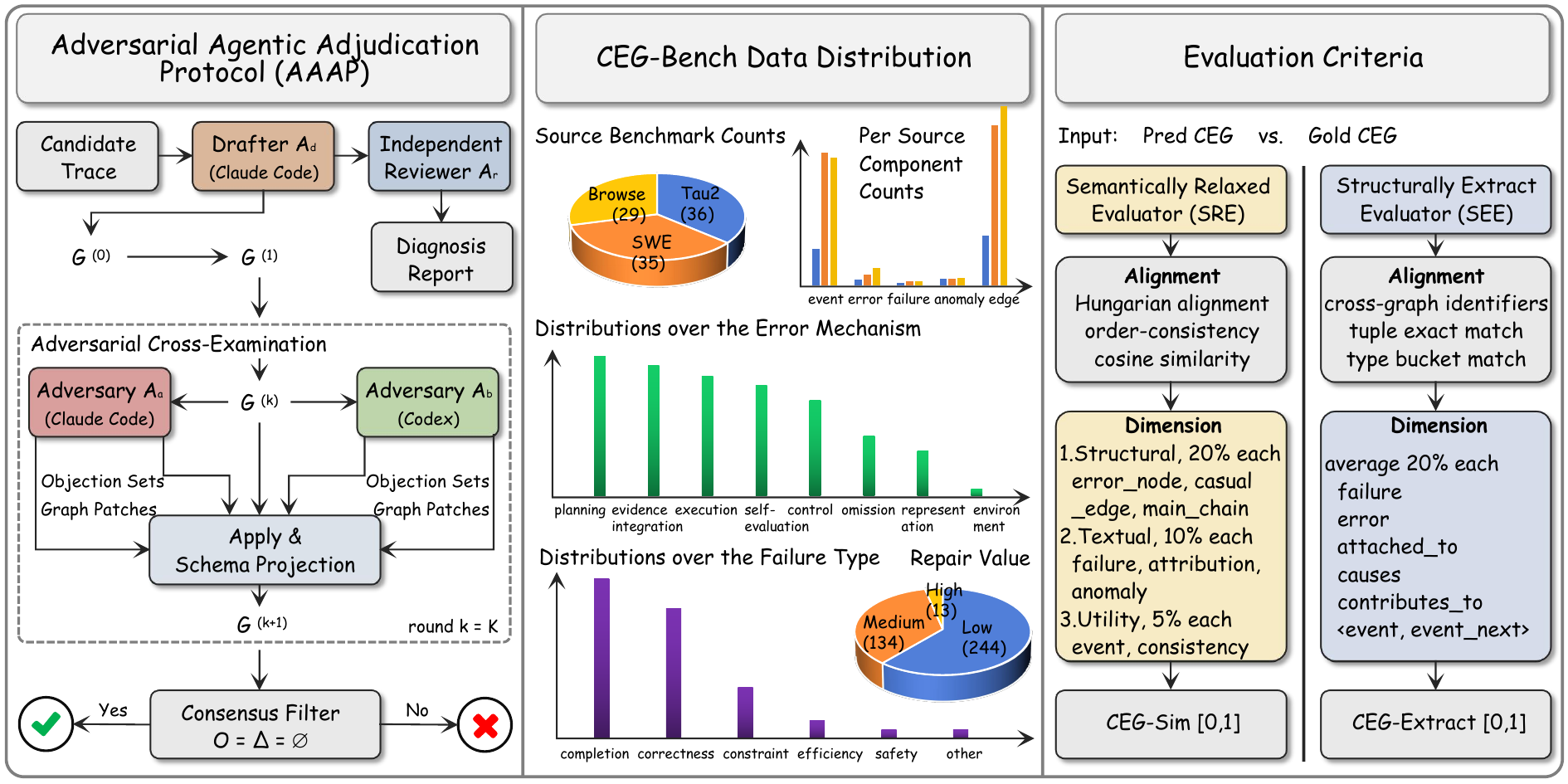}
  \vspace{-6mm}
  \caption{Construction and evaluation overview of CEG-Bench.}
  \label{fig3}
  \vspace{-4mm}
\end{figure*}

\subsubsection{Core Diagnosis Tools}

\textbf{Event construction} segments the normalized trace into ordered thought--action--observation units \citep{tracesir}, yielding the event set $V_E$ that provides the temporal scaffold against which every subsequent diagnostic node is grounded.

\textbf{Failure analysis} identifies the outcome-level failures $V_F$ against which the trace is to be explained, typically yielding a single primary failure $f^\star$, or at most three, which gives downstream error generation a concrete attribution target.

\textbf{Error hypothesis generation} proposes a compact set of causally relevant process-level errors $V_R$, each grounded in a specific event via $\varepsilon(r)$ and enriched with a mechanism $\mu(r)$, structural role $\rho(r)$, and trace-evidenced description $d(r)$. 

\textbf{Causal graph induction} adds the typed causal edges (\textsc{causes}, \textsc{amplifies}, \textsc{contributes-to}) that explain how errors propagate and accumulate into failures. Temporal (\textsc{event-next}) and grounding (\textsc{attached-to}) edges are populated deterministically from event order and from $\varepsilon(r)$.

\textbf{Anomaly detection} surfaces phenomenon-level irregularities $V_A$ that are observed but explicitly excluded from causal attribution. 
\textbf{Graph validation} checks $G_t$ against the well-formedness manifold $\mathcal{G}^\star(T)$ and returns an issue set $\mathcal{I}(G_t)$. Structurally repairable schema violations are handled by a deterministic projection operator $\Pi:\mathcal{G}\!\rightarrow\!\mathcal{G}^\star(T)$, while semantic inconsistencies are surfaced to an LLM critic, which emits a compact graph patch for iterative refinement. As a final safeguard, CEG-Agent projects the terminal graph $G_T$ to $G^\star=\Pi(G_T)$, ensuring that the released graph satisfies $\mathrm{W}_1\!\wedge\!\mathrm{W}_2\!\wedge\!\mathrm{W}_3$ regardless of policy realization. Further details on the checks, the $\Pi$, and the critic loop are provided in \S\ref{sec:impl-repair} and \S\ref{sec:impl-critic}.

\textbf{Repair-value estimation} assigns each error $r \in V_R$ a counterfactual repair value $\rho^\sharp(r) \in \{\textsc{high}, \textsc{medium}, \textsc{low}\}$ that scores whether fixing $r$ alone would prevent or substantially mitigate $f^\star$. The full counterfactual functional is given in \S\ref{sec:impl-repair-value}.


\subsubsection{Auxiliary Tools}

\textbf{Trace and state inspection} provides read-only access to the normalized trace and the current graph $G_t$, allowing the agent to re-examine evidence when an earlier decision needs to be revisited.

\textbf{Progress tracking} maintains a structured todo buffer inside $\mathcal{Q}_t$, giving the agent an explicit working memory for multi-step refinement, analogous to the TodoWrite mechanism in Claude Code.

\textbf{Sub-agent delegation} dispatches a focused read-only sub-agent to investigate a narrow question and returns a structured report that the parent agent may act on, while keeping the sub-agent isolated from the parent state. Further details are given in \S\ref{sec:impl-subagent}.

\textbf{Graph visualization} serializes the CEG into DOT, HTML, or SVG, enabling the agent and users to inspect error-propagation chains visually.

\textbf{Finalization} is the terminal action that exits the inference loop, after which the schema projection $\Pi$ is applied to guarantee a well-formed output.

\section{CEG-Bench}
\label{sec:benchmark}


\subsection{Trace Collection}
\label{sec:bench-data}

We construct \textbf{CEG-Bench}, a benchmark of 100 agentic execution traces annotated with consensus Causal Error Graphs. Following \citet{tracesir}, we sample 150 failed GLM-4.6 traces from three real-world agentic benchmarks: BrowseComp \citep{browsecomp-deepresearch} for deep research, Tau2Bench \citep{tau2-functioncalling} for function calling, and SWE-bench \citep{swebench-coding} for agentic coding, with 50 candidate traces from each source. All traces are kept in their original OpenAI-style message format, including assistant reasoning, tool calls, and observations. After AAAP, 100 consensus traces are retained for release: 29 from BrowseComp, 36 from Tau2Bench, and 35 from SWE-bench. 
The construction process is shown in Figure \ref{fig3}, and detailed analyses are provided in Appendix~\ref{app:bench-details}.



\subsection{Benchmark Construction}
\label{sec:bench-aaap}

The complexity of long agentic traces and the difficulty of manually identifying causal structure make conventional annotation workflows impractical. We therefore design an \textbf{Adversarial Agentic Adjudication Protocol} (AAAP), in which four heterogeneous agentic annotators play asymmetric roles, i.e., one drafter, one independent reviewer, and two adversarial cross-examiners. A trace is retained only if both adversaries certify the resulting CEG. This asymmetry is deliberate: the drafter--reviewer pair supports constructive graph proposal and refinement, while the two-adversary phase iteratively stress-tests and revises the CEG before imposing a strict consensus filter.

Let $\mathcal{A}_d$ denote the drafter, $\mathcal{A}_r$ the independent reviewer, and $\mathcal{A}_a, \mathcal{A}_b$ two cross-examining adversaries. We instantiate $\mathcal{A}_d, \mathcal{A}_a$ as Claude Code powered by Claude-Opus-4.7, $\mathcal{A}_r$ as the open-source TraceSIR analyzer powered by Claude-Opus-4.6 \citep{tracesir}, and $\mathcal{A}_b$ as Codex powered by GPT-5.5. The core principle of AAAP is that if a CEG is accepted by two adversarial agents built on frontier LLMs and state-of-the-art agentic coding frameworks, then it provides a credible consensus reference for causal trace diagnosis, despite the limitations of fully automated adjudication.


Specifically, AAAP proceeds in four stages:
(1) \textbf{Draft.} The drafter produces an initial graph $G^{(0)} = \mathcal{A}_d(T)$, conditioned only on $T$.
(2) \textbf{Independent review.} The reviewer independently analyzes the trace and emits a diagnostic report $\kappa^{(0)}=\mathcal{A}_r(T)$, identifying candidate errors from an additional perspective. The drafter incorporates this report to obtain $G^{(1)}=\mathcal{A}_d(T,G^{(0)},\kappa^{(0)})$.
(3) \textbf{Adversarial cross-examination.} For rounds $k=1,\ldots,K$, the two adversaries independently inspect and revise $G^{(k)}$, returning objection sets and graph patches $(O_a^{(k)},\Delta_a^{(k)})=\mathcal{A}_a(T,G^{(k)})$ and $(O_b^{(k)},\Delta_b^{(k)})=\mathcal{A}_b(T,G^{(k)})$. The next graph is obtained by applying the proposed patches and projecting back to the schema, $G^{(k+1)}=\Pi\bigl(\mathrm{Apply}(G^{(k)},\Delta_a^{(k)}\cup\Delta_b^{(k)})\bigr)$, where $\Pi$ is the schema projection.
(4) \textbf{Consensus filter.} The trace is retained iff there exists $k^\star \leq K$ such that $O_a^{(k^\star)}=O_b^{(k^\star)}=\varnothing$ and $\Delta_a^{(k^\star)}=\Delta_b^{(k^\star)}=\varnothing$. The accepted annotation is $G^\star=\Pi(G^{(k^\star)})$; otherwise, the trace is discarded. The retained traces therefore come with CEG annotations that have passed iterative adversarial revision and dual-adversary consensus, providing high-quality graphs.


\begin{wrapfigure}{r}{0.5\textwidth}
\centering
\vspace{-\intextsep}
\includegraphics[width=0.45\textwidth]{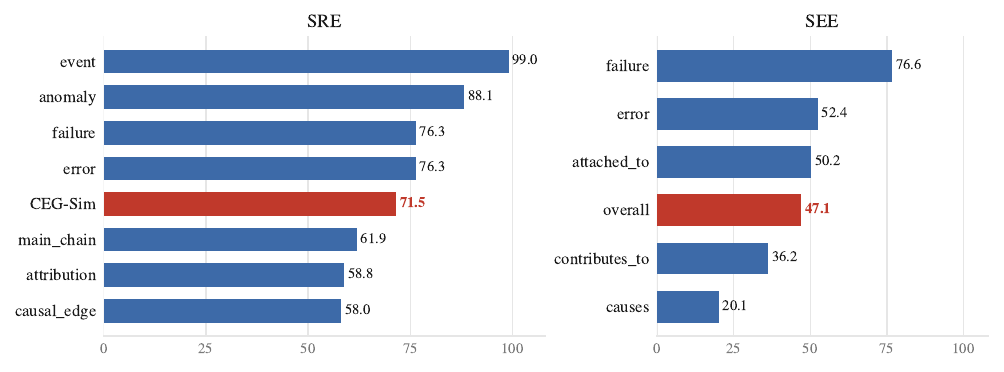}
\caption{AAAP--human-gold agreement per SRE dimension over all 100 traces.}
\label{fig:gold-similarity}
\end{wrapfigure}


\paragraph{Human validation.}
To verify the reliability of AAAP annotations, we additionally invited eight domain experts to independently cross-annotate all 100 traces, yielding a human gold set, \textbf{CEG-Bench-Gold}. When evaluated using the metrics in \S\ref{sec:bench-eval}, AAAP shows strong agreement with this human gold set, as illustrated in Figure~\ref{fig:gold-similarity}. Specifically, it achieves an overall $\mathrm{CEG\text{-}Sim}$ of $71.5$, with agreement scores of $76.3$, $76.3$, and $88.1$ on errors, failures, and anomalies, respectively. Achieving this level of agreement on such a challenging causal annotation task is non-trivial. More importantly, high-quality human annotation of long causal traces is prohibitively expensive and unavailable in most real deployments, whereas AAAP is fully automated and now validated against expert judgments. Producing scalable and reliable annotations is therefore itself a core contribution of this work. We adopt the AAAP-constructed \textbf{CEG-Bench} as the primary reference throughout the paper, while using \textbf{CEG-Bench-Gold} to \emph{audit} its reliability. The construction of CEG-Bench-Gold and per-system scores against the human gold set are provided in Appendix~\ref{app:human-gold}; all other experiments are reported against the AAAP annotations.

\newcolumntype{Y}{>{\centering\arraybackslash}X}
\newcommand{\totalcell}[1]{\cellcolor{blue!7}#1}
\newcommand{\besttotal}[1]{\cellcolor{blue!12}\textbf{#1}}

\begin{table*}[t]
\centering
\small
\setlength{\tabcolsep}{3.0pt}
\renewcommand{\arraystretch}{1.12}
\begin{tabularx}{\textwidth}{@{}
>{\raggedright\arraybackslash}p{2.35cm}
>{\raggedright\arraybackslash}p{3.25cm}
c
YYYYYYYY}
\toprule
\rowcolor{gray!12}
\textbf{Backbone} & \textbf{System}
& \cellcolor{blue!14}\textbf{SRE}
& \multicolumn{3}{c}{\textbf{Structural}}
& \multicolumn{3}{c}{\textbf{Textual}}
& \multicolumn{2}{c}{\textbf{Utility}} \\
\rowcolor{gray!12}
&
&
& \texttt{error}
& \texttt{edge}
& \texttt{chain}
& \texttt{fail.}
& \texttt{attr.}
& \texttt{anom.}
& \texttt{event}
& \texttt{cons.} \\
\cmidrule(lr){3-3}
\cmidrule(lr){4-6}
\cmidrule(lr){7-9}
\cmidrule(l){10-11}
Claude-Opus-4.7
& Claude Code (raw)
& \totalcell{45.8}
& 39.5 & 37.1 & 38.8 & 44.7 & \textbf{48.7} & 44.7 & 78.1 & 100.0 \\
Claude-Opus-4.7
& Claude Code (+Process)
& \totalcell{49.6}
& \textbf{49.2} & 41.8 & 38.2 & \textbf{47.7} & 43.9 & 46.0 & 100.0 & 100.0 \\
\rowcolor{blue!3}
Claude-Opus-4.7
& \textbf{CEG-Claude} (ours)
& \besttotal{51.6}
& 46.7 & \textbf{42.4} & \textbf{51.0} & 38.8 & 37.1 & \textbf{60.0} & 100.0 & 99.9 \\
\addlinespace[2pt]
\midrule
GPT-5.5
& Codex (raw)
& \totalcell{32.0}
& 22.8 & 31.8 & 27.4 & 29.9 & 25.5 & 2.0 & 98.0 & 100.0 \\
GPT-5.5
& Codex (+Process)
& \totalcell{32.9}
& 27.0 & 34.5 & 29.5 & 21.5 & 23.5 & 1.7 & 100.0 & 100.0 \\
\rowcolor{blue!3}
GPT-5.5
& \textbf{CEG-Codex} (ours)
& \besttotal{46.5}
& \textbf{45.9} & \textbf{38.0} & \textbf{41.5} & \textbf{34.9} & \textbf{27.6} & \textbf{51.3} & 100.0 & 100.0 \\
\bottomrule
\end{tabularx}
\vspace{-3mm}
\caption{Semantically relaxed evaluation (\textbf{SRE}) on CEG-Bench.}
\label{tab:exp-sre}
\vspace{-3mm}
\end{table*}

\begin{table*}[t]
\centering
\small
\setlength{\tabcolsep}{4.0pt}
\renewcommand{\arraystretch}{1.12}
\begin{tabularx}{\textwidth}{@{}
>{\raggedright\arraybackslash}p{2.35cm}
>{\raggedright\arraybackslash}p{3.25cm}
c
YYYYY}
\toprule
\rowcolor{gray!12}
\textbf{Backbone} & \textbf{System}
& \cellcolor{blue!14}\textbf{SEE}
& \multicolumn{5}{c}{\textbf{Dimensions}} \\
\rowcolor{gray!12}
&
&
& \texttt{failure}
& \texttt{error}
& \texttt{attach}
& \texttt{cause}
& \texttt{contrib.} \\
\cmidrule(lr){3-3}
\cmidrule(l){4-8}
Claude-Opus-4.7
& Claude Code (raw)
& \totalcell{13.9}
& \textbf{56.1} & 5.3 & 3.9 & 1.0 & 3.3 \\
Claude-Opus-4.7
& Claude Code (+Process)
& \totalcell{19.3}
& 47.3 & 18.5 & 16.3 & 1.1 & \textbf{13.2} \\
\rowcolor{blue!3}
Claude-Opus-4.7
& \textbf{CEG-Claude} (ours)
& \besttotal{20.5}
& 52.1 & \textbf{20.5} & \textbf{17.8} & \textbf{2.0} & 9.9 \\
\addlinespace[2pt]
\midrule
GPT-5.5
& Codex (raw)
& \totalcell{11.7}
& 43.8 & 5.6 & 6.0 & 0.0 & 3.1 \\
GPT-5.5
& Codex (+Process)
& \totalcell{10.4}
& 33.3 & 8.7 & 6.0 & 0.0 & 4.1 \\
\rowcolor{blue!3}
GPT-5.5
& \textbf{CEG-Codex} (ours)
& \besttotal{16.9}
& \textbf{48.3} & \textbf{15.2} & \textbf{13.0} & \textbf{1.5} & \textbf{6.3} \\
\bottomrule
\end{tabularx}
\vspace{-3mm}
\caption{Structurally exact evaluation (\textbf{SEE}) on CEG-Bench.}
\label{tab:exp-see}
\vspace{-3mm}
\end{table*}


\subsection{Evaluation Criteria}
\label{sec:bench-eval}

A predicted CEG $\hat{G}$ is evaluated against the consensus annotation $G^\star \in \mathcal{G}^\star(T)$ using two complementary criteria. The \textbf{semantically relaxed} evaluator (SRE) aligns predicted and gold nodes via Hungarian bipartite matching, using a type-specific similarity that combines event correspondence, categorical agreement on $\mu(r), \rho(r), \theta(f)$, and text similarity \citep{bge-m3} over $\nu(r), d(r)$. 
Here, $\mu(r)$, $\rho(r)$, and $\theta(f)$ are controlled categorical fields with 8, 3, and 6 predefined values, respectively, as shown in Figure \ref{fig3}, whereas $\nu(r)$ and $d(r)$ are free-text fields generated by the LLM.
After alignment, predicted identifiers are translated into the gold space, and each dimension is scored by precision (P), recall (R), and F1. The resulting $\mathrm{CEG\text{-}Sim}(\hat{G},G^\star)\in[0,1]$ is a tier-weighted aggregate over three categories: \textbf{structural} dimensions, \texttt{error\_node}, \texttt{causal\_edge}, and \texttt{main\_chain}, weighted by $20\%$ each; \textbf{textual} dimensions, \texttt{failure}, \texttt{attribution}, and \texttt{anomaly}, weighted by $10\%$ each; and \textbf{utility} dimensions, \texttt{event} and \texttt{consistency}, weighted by $5\%$ each.
The \textbf{structurally exact} evaluator (SEE) performs no fuzzy matching, resolving errors by the exact key $(\varepsilon(r),\mu(r))$ and failures by the bucket $\theta(f)$. Edges are compared as multisets of typed tuples after translating predicted identifiers into the gold space, with edges incident to unmatched nodes counted as false positives. The resulting $\mathrm{CEG\text{-}Exact}(\hat{G},G^\star)\in[0,1]$ is the mean of five discriminative dimensions: \texttt{failure}, \texttt{error}, \texttt{attached\_to}, \texttt{causes}, and \texttt{contributes\_to}. Details are provided in Appendix~\ref{app:eval-spec}.

\section{Experiments}
\label{sec:exp}

We evaluate \textbf{CEG-Agent} on \textbf{CEG-Bench}. For each backbone family, we compare three systems on the same 100 traces: a raw coding-agent baseline given the original trace and CEG schema (\textbf{raw}), a stronger baseline given deterministic event segmentation $V_E$ and \textsc{event-next} edges (\textbf{+Process}), and \textbf{CEG-Agent} instantiated with the same backbone. We use Claude-Opus-4.7 for the Claude Code / CEG-Claude comparison and GPT-5.5 for the Codex / CEG-Codex comparison. Because no existing system, to our knowledge, emits typed causal error graphs under our schema, a strictly like-for-like external baseline does not yet exist; adapting a flat-classification or step-scoring method would itself require the schema-projection scaffolding we contribute. Our \textbf{+Process} baseline instead reproduces the deterministic event-segmentation stage that prior trace-diagnosis methods rely on. We report \textbf{SRE} ($\mathrm{CEG\text{-}Sim}$) and \textbf{SEE} ($\mathrm{CEG\text{-}Exact}$) from \S\ref{sec:bench-eval}; higher is better, and all scores are percentages. CEG-Agent uses inference budget $K=24$, critic budget $C=2$, and temperature $=1$, while baselines use their CLI defaults. Detailed experimental analyses are provided in Appendix~\ref{app:exp-details}.



\subsection{Overall Performance}
\label{sec:exp-overall}

Tables~\ref{tab:exp-sre} and \ref{tab:exp-see} report the headline results. \textbf{CEG-Agent achieves the best total score under both SRE and SEE for both backbone families.} With Claude-Opus-4.7, CEG-Claude improves over the stronger Claude Code (+Process) baseline by $+2.0$ points in SRE and $+1.2$ points in SEE. With GPT-5.5, the gains are larger, i.e., CEG-Codex improves over Codex (+Process) by $+13.6$ points in SRE and $+6.5$ points in SEE.
These gains are not merely due to better event parsing, since the +Process baselines already receive the deterministic event scaffold. The improvements therefore mainly reflect better causal diagnosis, graph construction, and refinement. The benefit is especially pronounced for GPT-5.5, suggesting that structured tool use and validation can compensate for weaker raw diagnostic behavior. 
Overall, CEG-Agent's anomaly--error--failure taxonomy and typed causal graph construction make long trace diagnosis more actionable and structurally faithful.

\subsection{Fine-Grained Analysis}
\label{sec:exp-synthesis}

The per-dimension results in Tables~\ref{tab:exp-sre} and \ref{tab:exp-see} localize the main sources of improvement. Both CEG-Agent variants lead their same-backbone baselines on the core structural dimensions, with the largest gains on SRE \texttt{main\_chain}, \texttt{causal\_edge}, \texttt{anomaly}, and SEE \texttt{error}, \texttt{attached\_to}, and \texttt{causes}. The few apparent exceptions under Claude, including SRE \texttt{error}/\texttt{failure}/\texttt{attribution} and SEE \texttt{contributes\_to}, are mainly due to gold-empty traces, where abstaining systems are mechanically favored, and textual-style mismatch between detailed predictions and compact gold descriptions (\S\ref{app:exp-empty}, \S\ref{app:exp-dim}). On the non-empty subset, CEG-Agent leads same-backbone baselines by $+5.6$/$+1.4$ points under Claude and $+13.0$/$+6.2$ points under GPT-5.5 on SRE/SEE, respectively, suggesting that the headline results understate its advantage on substantive causal diagnosis.
Additional analyses in \S\ref{app:exp-details} explain these gains. First, CEG-Agent better matches the scale of the gold graphs, producing $\bar{|V_R|}\approx4.6$ errors and $\bar{|E_C|}\approx6.9$ causal edges, close to the gold averages of $3.9$ and $7.5$, while baselines under-generate errors, anomalies, and causal edges (\S\ref{app:exp-calibration}). Second, its largest recall gains occur on difficult mechanisms such as \emph{representation}, \emph{evidence integration}, and \emph{planning}, as well as on \textsc{propagated} errors that require causal-chain reconstruction (\S\ref{app:exp-mech-role}). Third, the advantage is concentrated on medium-to-large traces that require sustained causal reconstruction, where incremental graph construction is most useful (\S\ref{app:exp-complexity}). Repair-value calibration, cost and tool use, and visualization case studies are analyzed in \S\ref{app:exp-repair}, \S\ref{app:exp-cost}, and \S\ref{app:exp-case}, respectively.
Moreover, CEG construction remains challenging. Even the best system reaches only $51.6$ SRE and $20.5$ SEE, indicating strong agents still struggle to identify causally relevant errors and reconstruct their propagation chains.

\subsection{Cross-Backbone Generalization}
\label{sec:exp-backbone}

Table~\ref{tab:exp-backbone} evaluates CEG-Agent with five LLM backbones under the same schema, tools, and budgets. CEG-Agent produces useful diagnostic graphs across all tested models: even the weakest SRE backbone, Qwen3.7-Max, reaches $43.7$, well above the strongest Codex baseline at $32.9$. This suggests that the validator-guided construction loop provides a backbone-agnostic structural scaffold.

\begin{wraptable}{r}{0.42\textwidth}
\centering
\vspace{-\intextsep}
\small
\setlength{\tabcolsep}{5pt}
\renewcommand{\arraystretch}{1.05}
\begin{tabular}{lcc}
\toprule
\textbf{Backbone} & \textbf{SRE} & \textbf{SEE} \\
\midrule
Claude-Opus-4.7 & \textbf{51.6} & \textbf{20.5} \\
GLM-5.1         & 46.7 & 19.6 \\
GPT-5.5         & 46.5 & 16.9 \\
Gemini-3.1-Pro  & 44.2 & 15.0 \\
Qwen3.7-Max     & 43.7 & 18.7 \\
\bottomrule
\end{tabular}
\caption{CEG-Agent composite SRE/SEE under five LLM backbones.}
\label{tab:exp-backbone}
\vspace{-0.4cm}
\end{wraptable}


The SRE and SEE rankings also diverge beyond the top model. In particular, Qwen3.7-Max outperforms GPT-5.5 and Gemini-3.1-Pro on structurally exact reconstruction despite lower semantic similarity, indicating that fluent diagnostic descriptions and exact causal structure are distinct capabilities. The fixed framework further exposes backbone-specific profiles, such as Claude's overall strength, GLM-5.1's strong exact reconstruction, GPT-5.5's better semantic than structural behavior, Gemini's conservative contribution edges, and Qwen's weaker anomaly coverage. Together, these results show that CEG-Agent is not tied to a single LLM family and that CEG evaluation can serve as a discriminative probe of causal-diagnosis capability.

\subsection{Closed-Loop Repair with CEG Feedback}
\label{sec:exp-optimization}

Beyond diagnostic quality, we also probe whether CEGs can serve as actionable repair signals. Since all CEG-Bench traces are failed attempts, the original runs have zero success by construction. We re-attempt each task with a feedback hint derived from the AAAP CEG, comparing a content-free {generic} retry note with three compact CEG-based hints: the outcome-level failure, the causal {errors}, and {priority errors}, i.e., errors with high or medium repair value. 
Details are provided in Appendix~\ref{app:exp-optimization}.

\begin{wraptable}{r}{0.43\textwidth}
\vspace{-\intextsep}
\small
\centering
\setlength{\tabcolsep}{4pt}
\renewcommand{\arraystretch}{1.05}
\begin{tabular}{lccc}
\toprule
\textbf{Feedback} & \textbf{Tau2} & \textbf{Browse} & \textbf{SWE} \\
\midrule
generic              & 77.8 & 6.9  & 14.3 \\
\midrule
CEG: failure         & \textbf{91.7} & 10.3 & \textbf{17.1} \\
CEG: errors          & 80.6 & \textbf{17.2} & \textbf{17.1} \\
CEG: priority errors & 83.3 & \textbf{17.2} & 8.6 \\
\bottomrule
\end{tabular}
\vspace{-3mm}
\caption{Closed-loop repair pass@1 (\%).}
\label{tab:exp-opt}
\vspace{-0.4cm}
\end{wraptable}

Table~\ref{tab:exp-opt} shows that CEG is most effective when agent can act on it interactively. For example, on Tau2Bench, all CEG-derived hints outperform the generic retry, with the failure-level hint giving the largest gain. This suggests that explicitly naming the outcome-level failure provides a concrete correction target, while causal errors and repair-prioritized errors offer useful process-level guidance. The fact that structured CEG hints consistently win further indicates that the gains directly come from the diagnostic content of CEG.

The preferred granularity of feedback depends on the task. On BrowseComp, error-based hints are more useful than the failure-only hint, likely because the task requires revisiting evidence and reasoning steps rather than merely knowing that the final answer is wrong. However, the absolute success rate remains low, since many failed traces do not contain enough missing evidence for one-shot re-derivation. On SWE-bench, concise failure or error hints slightly improve over generic feedback, whereas priority-error filtering hurts. This suggests that repair values are not yet sufficiently calibrated for single-shot code patching: filtering to high-value or medium-value errors may omit contextual errors needed for a correct patch, or over-focus the model on an incomplete local fix.
Overall, these results should be viewed as directional rather than conclusive, given the small per-source sample sizes. Still, they show that CEGs can provide useful repair signals beyond diagnosis, especially in interactive settings. We leave larger-scale closed-loop optimization to future work.

\section{Conclusion}
\label{sec:conclusion}


We presented \textbf{CEG-Agent}, a tool-augmented agentic framework for causal diagnosis of agentic execution traces. Instead of treating all diagnostic signals as generic errors, CEG-Agent separates anomalies, errors, and failures, and represents their relationships using \textbf{Causal Error Graphs} (CEGs), a typed graph structure that connects trace events, process-level causal errors, and outcome-level failures. To support systematic evaluation, we further introduced \textbf{CEG-Bench}, a benchmark of agentic traces annotated with consensus CEGs through an \textbf{Adversarial Agentic Adjudication Protocol}, and validated its annotations against the expert-curated \textbf{CEG-Bench-Gold}. Experiments under both \textbf{semantically relaxed} and \textbf{structurally exact} evaluation criteria show that CEG-Agent outperforms strong agentic baselines in constructing causal error graphs. By producing visualizable propagation structures and repair-oriented recommendations, CEG-Agent provides actionable diagnostic evidence for debugging agent executions and improving future agent behavior.

\subsection*{AI Use Statement}

In this work, we used generative AI tools for {generating AI-assisted dataset annotations}: the CEG annotations in \textbf{CEG-Bench} are constructed by agentic LLM systems through the Adversarial Agentic Adjudication Protocol (AAAP, \S\ref{sec:bench-aaap}), in which frontier LLMs draft, independently review, and adversarially cross-examine each Causal Error Graph annotation until dual-adversary consensus is reached. These AI-generated annotations, together with the deterministic repair values derived from them, are central artifacts of this paper. The CEG-Agent diagnoses and the closed-loop repair study are likewise produced by LLM-driven agents, as described in the corresponding sections.
Additionally, we used generative AI tools {to aid and polish the writing} of the manuscript, improving grammar and clarity of author-drafted text. We did {not} use generative AI tools to prove mathematical claims or for the core research ideation of this work, and generative-AI-assisted retrieval or discovery of related work is not applicable here.
We reviewed all AI-assisted outputs. The AI-generated CEG-Bench annotations were validated against an independent expert-curated human gold set, \textbf{CEG-Bench-Gold} (\S\ref{app:human-gold}), and we report and analyze their close agreement with this human reference. All quantitative results were re-checked by the authors against the released evaluation code, and all AI-polished text was verified for accuracy and faithfulness to the intended meaning. We take responsibility for the final content of this work, including all text, claims, and artifacts produced with the aid of AI.

\bibliographystyle{iclr2027_conference}
\bibliography{iclr2027_conference}

\appendix

\begin{table*}[t]
\small
\centering
\renewcommand{\arraystretch}{1.12}
\setlength{\tabcolsep}{4pt}
\begin{tabularx}{\textwidth}{@{}p{1.6cm}p{3.2cm}p{2.2cm}p{1.6cm}X@{}}
\toprule
\rowcolor{gray!18}
\textbf{Category} & \textbf{Tool} & \textbf{Method} & \textbf{Mode} & \textbf{Effect on $S_t$} \\
\midrule
Inspection
& \texttt{read\_trace}
& Trace inspection
& read-only
& Read the normalized trace $T$. \\

Inspection
& \texttt{inspect\_state}
& State inspection
& read-only
& Read the current graph $G_t$ and todo buffer $\mathcal{Q}_t$. \\

\midrule
Phase
& \texttt{build\_events}
& Event construction
& generator
& Write event nodes $V_E$. \\

Phase
& \texttt{find\_failures}
& Failure analysis
& generator
& Write failure nodes $V_F$. \\

Phase
& \texttt{propose\_errors}
& Error hypothesis generation
& generator
& Write causally relevant error nodes $V_R$. \\

Phase
& \texttt{build\_causal\_graph}
& Causal graph induction
& generator
& Write causal edges $E_C$. \\

Phase
& \texttt{find\_anomalies}
& Anomaly detection
& generator
& Write anomaly nodes $V_A$. \\

Phase
& \texttt{estimate\_repair\_values}
& Repair-value estimation
& optional
& Write repair values $\rho^\sharp$. \\

\midrule
Critic
& \texttt{validate\_graph}
& Graph validation and schema projection
& deterministic
& Emit issue set $\mathcal{I}(G_t)$ and apply $G_t\leftarrow\Pi(G_t)$ to structurally repairable violations. \\

Critic
& \texttt{apply\_critic\_patch}
& Patch-based critic loop
& critic patch
& Apply an LLM critic patch $p$ and project back to the schema: $G_t\leftarrow\Pi(\mathrm{Apply}(G_t,p))$. \\

\midrule
Meta
& \texttt{write\_todo}
& Progress tracking
& state update
& Update the todo buffer $\mathcal{Q}_t$. \\

Meta
& \texttt{spawn\_subagent}
& Sub-agent delegation
& delegation
& Dispatch a read-only sub-agent for localized investigation. \\

Meta
& \texttt{render\_graph}
& Graph visualization
& serialization
& Export the CEG as DOT, HTML, or SVG. \\

\midrule
Terminal
& \texttt{finalize}
& Finalization
& terminal
& Halt the inference loop and trigger final projection. \\
\bottomrule
\end{tabularx}
\caption{Registered tool set $\mathcal{U}$ of CEG-Agent. Tools are grouped by category, linked to the corresponding method component, and summarized by their execution mode and effect on the diagnosis state $S_t$.}
\label{tab:tools}
\end{table*}

\section{CEG-Agent Implementation Details}
\label{sec:impl}

This section instantiates the abstractions in \S\ref{sec:method} as the operational components of CEG-Agent. We describe trace normalization, the registered tool set and its dispatch table, the agentic runtime, the deterministic validator and auto-repair rules that realize the schema projection $\Pi$, the LLM-based critic loop with its compact patch language, the counterfactual repair-value functional, the sub-agent delegation interface, and the per-trace output bundle.

\subsection{Trace Normalization}
\label{sec:impl-trace}

A deterministic preprocessor maps each provider-specific message $m_i\in\mathcal{M}$ to a normalized turn. Assistant messages are decomposed into a \textit{thought} field, extracted from \texttt{reasoning\_content} when available and otherwise from \texttt{content}, and an \textit{action} field, extracted from serialized \texttt{tool\_calls} when present. Tool messages, as well as user messages that serve as environment or human responses to a preceding assistant action, are attached as the \textit{observation} of the corresponding assistant turn, yielding a thought--action--observation triple for each execution unit. The first user message is treated as the task specification $\tau$, while subsequent user turns are retained when they provide feedback or observations during the interaction. The normalized trace preserves the original execution order and is stored in the output bundle for auditability.

\subsection{Tool Set}
\label{sec:impl-tools}

The registered tool set $\mathcal{U}$ contains fourteen tools, organized by their role in the diagnosis state $S_t$ and by the method component they implement, as shown in Table~\ref{tab:tools}. Each tool is exposed through an OpenAI-style function-calling JSON schema and dispatched by name. Core diagnosis tools update specific components of the CEG, while inspection and meta tools provide read-only access, working-memory management, sub-agent delegation, and visualization. The repair-value tool is enabled by default and can be disabled through \texttt{CE\_ENABLE\_REPAIR}; when disabled, \texttt{estimate\_repair\_values} is not registered and therefore cannot be invoked by the agent. The number of critic-based refinement rounds is controlled by \texttt{CE\_CRITIC\_ROUNDS}. At runtime, CEG-Agent automatically selects and invokes these tools based on the evolving diagnosis state and refinement needs.

\subsection{Agentic Runtime}
\label{sec:impl-runtime}

\begin{algorithm}[t]
\small
\caption{CEG-Agent Inference}
\label{alg:ceg-agent}
\begin{algorithmic}[1]
\Require trace $T$, tool set $\mathcal{U}$, budget $K$
\State $S_0\!\gets\!\textsc{InitState}(T)$;\; $G_0\!\gets\!\varnothing$
\For{$t = 0, \ldots, K{-}1$}
  \State $(u_t,\mathbf{a}_t) \sim \pi_\theta(\,\cdot\mid S_t,\mathcal{U})$
  \State $S_{t+1} \gets \Phi_{u_t}(S_t,\mathbf{a}_t)$
  \State \textbf{if} $u_t = \texttt{finalize}$ \textbf{then break}
\EndFor
\State \Return $G^\star \gets \Pi(G_{t+1})$ \Comment{schema projection}
\end{algorithmic}
\end{algorithm}

The runtime implements the inference loop in Algorithm~\ref{alg:ceg-agent}. At each iteration, the top-level LLM receives the conversation history $H_t$, a JSON serialization of the current diagnosis state $S_t$, and the tool specifications in $\mathcal{U}$. It then emits zero or more tool calls, which are dispatched sequentially and applied to the state through the corresponding transition functions $\Phi_u$. The loop terminates when the model invokes \texttt{finalize}, emits no tool call, or reaches the iteration budget, which is set to $K=24$ in our experiments.

\paragraph{Per-tool exception isolation.}
If a tool handler raises an exception or returns malformed output, the runtime intercepts the failure and returns a structured error object $\{\texttt{error}:\cdot\}$ as the tool result, while leaving the graph state $G_t$ unchanged. This allows the policy to recover from local tool failures, such as malformed LLM-generated JSON inside a phase tool, without aborting the entire trajectory.

\paragraph{Termination safety net.}
After the loop exits, the schema projection $\Pi$ is invoked once more regardless of whether the model explicitly called \texttt{finalize}. This final projection ensures that the released graph $G^\star=\Pi(G_t)$ always conforms to the CEG schema and satisfies the required well-formedness constraints.

\subsection{Schema Projection}
\label{sec:impl-repair}

CEG inference can produce two qualitatively different types of issues. \emph{Structural} violations, such as illegal edge types, dangling endpoints, duplicate edges, missing schema edges, and role--topology mismatches, can be repaired mechanically. In contrast, \emph{semantic} violations, such as duplicated failures, anomaly--error confusion, or evidence-dependent role errors, require trace-level judgment. CEG-Agent separates these cases using a deterministic validator together with an LLM-driven patch critic (\S\ref{sec:impl-critic}).

\paragraph{Validator and issue partition.}
Let $\mathcal{V}:\mathcal{G}\!\rightarrow\!2^{\mathrm{IssueCode}}$ be a deterministic validator that maps a graph $G$ to an issue set
$\mathcal{I}(G)=\mathcal{I}_{\mathrm{det}}(G)\,\dot{\cup}\,\mathcal{I}_{\mathrm{sem}}(G)$,
where the partition is determined by whether an issue admits a mechanical repair. The deterministic partition $\mathcal{I}_{\mathrm{det}}$ includes unknown edge types, illegal or dangling endpoints, duplicate edges, missing \textsc{event-next} chains, missing \textsc{attached-to} edges, and role--topology mismatches. The semantic partition $\mathcal{I}_{\mathrm{sem}}$ includes anomaly--error overlap, missing or excessive failures, errors unreachable from any failure, and role assignments that require trace evidence to resolve.

\paragraph{Deterministic auto-repair $\mathcal{R}_{\mathrm{det}}$.}
The auto-repair routine realizes the schema projection $\Pi$ through five deterministic rules: (i) remove edges that violate Eq.~\eqref{eq:typed-edge} or reference absent endpoints; (ii) deduplicate edges by $(\textit{source},\,\textit{type},\,\textit{target})$; (iii) regenerate the \textsc{event-next} chain from event order; (iv) regenerate \textsc{attached-to} edges from the event grounding $\varepsilon(r)$ of each error; and (v) recompute the structural role $\rho(r)$ from graph topology, demoting a \textsc{root} with incoming \textsc{causes} edges to \textsc{propagated} and assigning \textsc{amplification} when $r$ participates in any \textsc{amplifies} edge. These rules remove all mechanically repairable violations without relying on trace semantics, yielding:
\begin{equation}
\Pi(G) \;=\; \lim_{k\rightarrow\infty}\mathcal{R}_{\mathrm{det}}^{k}(G),
\quad
\mathcal{I}_{\mathrm{det}}\bigl(\Pi(G)\bigr)=\varnothing.
\label{eq:fixed-point}
\end{equation}
By construction, $\Pi$ is idempotent, i.e., $\Pi\circ\Pi=\Pi$. This allows the runtime to invoke schema projection both inside critic-based repair tools and once again at termination, ensuring that the released graph remains structurally well formed.

\subsection{LLM Critic Loop}
\label{sec:impl-critic}

Semantic issues in $\mathcal{I}_{\mathrm{sem}}(G)$ require evidence-based judgment and cannot be resolved by deterministic projection alone. CEG-Agent therefore uses an LLM critic policy $\pi_\theta^{\mathrm{crit}}$ to propose a compact \emph{graph patch} $p\in\mathcal{P}$, rather than regenerating the entire graph from scratch.

\paragraph{Patch language.}
A patch is represented as:
\begin{equation}
p = \bigl(\Delta^+_E,\,\Delta^-_E,\,\Delta^-_R,\,\Delta^-_F,\,\Delta^-_A,\,\Delta_{\mathrm{attr}},\,\Delta_{\mathrm{merge}}\bigr),
\label{eq:patch}
\end{equation}
where $\Delta^+_E$ and $\Delta^-_E$ denote edge insertions and deletions, $\Delta^-_R,\Delta^-_F,\Delta^-_A$ denote deletions of error, failure, and anomaly nodes, $\Delta_{\mathrm{attr}}$ denotes in-place attribute updates such as role, mechanism, or description edits, and the merge set $\Delta_{\mathrm{merge}}\subset V_R\times 2^{V_R}$ specifies error-node merges. The interpreter $\mathrm{Apply}(\cdot,p)$ executes these patch operations, including merges, deletions, updates, and insertions, and discards any operation that references an absent node or violates the typed-edge constraints in Eq.~\eqref{eq:typed-edge}.

\paragraph{Iterative refinement.}
Each critic round applies one patch and then projects the result back to the CEG schema:
\begin{equation}
\begin{aligned}
p^{(k)} &\sim \pi_\theta^{\mathrm{crit}}
    \bigl(\cdot \mid G^{(k)}, \mathcal{I}(G^{(k)})\bigr),\\
G^{(k+1)} &= \Pi\!\left(
    \mathrm{Apply}\bigl(G^{(k)},p^{(k)}\bigr)
\right).
\end{aligned}
\label{eq:critic-update}
\end{equation}
The number of critic rounds is bounded by a critic budget $C$, with default $C=2$. Since $\Pi$ is applied after every patch and once again at termination, the final graph $G^\star=\Pi(G^{(C)})$ satisfies $\mathrm{W}_1\!\wedge\!\mathrm{W}_2\!\wedge\!\mathrm{W}_3$ even when the critic proposes malformed edits. Restricting the critic to local patches makes the repair process auditable and limits the potential damage of any single revision.

\subsection{Counterfactual Repair Valuation}
\label{sec:impl-repair-value}

For each error $r\in V_R$ in the CEG, the tool \texttt{estimate\_repair\_values} assigns a counterfactual repair value
$\rho^\sharp(r)\in\{\textsc{high},\textsc{medium},\textsc{low}\}$, estimating whether correcting only $r$ would avoid or substantially mitigate the primary failure $f^\star$. The valuation combines four graph-structural signals with a bounded lexical signal in a clipped linear functional.

\paragraph{Structural primitives.}
Let $G_{-r}$ denote the induced subgraph obtained by deleting $r$ and all incident edges from $G$. We define the set of errors contributing to $f^\star$ as:
\begin{equation}
\mathcal{C}_G(f^\star)
=
\bigl\{q\in V_R \mid f^\star \in \mathrm{Reach}_G(q)\bigr\}.
\label{eq:contrib}
\end{equation}
The \emph{softened counterfactual severance} of $r$ is the fraction of contributing errors whose reachability to $f^\star$ is cut after removing $r$:
\begin{equation}
\begin{aligned}
\sigma(r,f^\star)
=
\frac{
\bigl|
\{q\in\mathcal{C}_G(f^\star)\setminus\{r\}
:
f^\star\notin \mathrm{Reach}_{G_{-r}}(q)\}
\bigr|
}{
|\mathcal{C}_G(f^\star)|
}.
\end{aligned}
\label{eq:severance}
\end{equation}
We set $\sigma(r,f^\star)=0$ when $r\notin\mathcal{C}_G(f^\star)$ or $\mathcal{C}_G(f^\star)=\varnothing$. A value $\sigma=1$ indicates that $r$ is a unique bottleneck, whereas smaller values indicate that $r$ is one of several parallel contributors.
It is worth noting that this graded computation is used for the gold annotations in CEG-Bench. In CEG-Agent inference, for simplicity, we ask the LLM to provide a binary estimate of this signal, i.e., whether $r$ acts as a counterfactual bottleneck or not.

\paragraph{Amplification credit.}
Let
$\mathrm{Succ}_{\textsc{ampl}}(r)=\{q:(r,\textsc{amplifies},q)\in E\}$.
The amplification credit is:
\begin{equation}
\alpha(r)
=
1-\exp\;\!\bigl(-|\mathrm{Succ}_{\textsc{ampl}}(r)|\bigr),
\label{eq:amp}
\end{equation}
a saturating function of the \textsc{amplifies} out-degree. This prevents an error from dominating the score solely by amplifying many downstream errors.

\paragraph{Softened upstream hazard.}
For propagated errors, we penalize cases where the upstream cause remains unresolved. Let
$\mathrm{Pred}_{\textsc{cause}}(r)=\{u:(u,\textsc{causes},r)\in E\}$.
The softened upstream hazard is:
\begin{equation}
\begin{aligned}
\eta(r)
&=
\mathbb{1}\!\bigl[\rho(r)=\textsc{propagated}\bigr]
\cdot
\frac{1}{|\mathrm{Pred}_{\textsc{cause}}(r)|}
\\
&\quad \cdot
\sum_{u\in\mathrm{Pred}_{\textsc{cause}}(r)}
\bigl(1-\sigma(u,f^\star)\bigr),
\end{aligned}
\label{eq:hazard}
\end{equation}
with the convention $\eta(r)=0$ when $\mathrm{Pred}_{\textsc{cause}}(r)=\varnothing$. Intuitively, $\eta(r)$ is small when upstream causes are themselves strong bottlenecks, and large when they are weak parallel contributors that would remain unresolved by fixing $r$ alone.
As with $\sigma$, this graded form is used for gold annotations in CEG-Bench, while CEG-Agent inference uses a binary surrogate $\eta(r) \in \{0,1\}$ that triggers when $r$ is propagated and has at least one upstream \textsc{causes} edge.

\paragraph{Composite score and bucketing.}
Let $\pi(\rho)$ be a role prior with
$\pi(\textsc{root})=1.0$,
$\pi(\textsc{amplification})=0.5$, and
$\pi(\textsc{propagated})=0.3$.
Let $h(r)\in[0.3,0.7]$ be a bounded lexical hedging signal extracted from $d(r)$. The composite repair score is:
\begin{equation}
\begin{aligned}
s(r)
=
\mathrm{clip}_{[0,1]}\!\Bigl(
& w_\sigma \sigma(r,f^\star)
+ w_\pi \pi(\rho(r))
+ w_\alpha \alpha(r)
\\
& + w_h h(r)
- w_\eta \eta(r)
\Bigr),
\end{aligned}
\label{eq:score}
\end{equation}
with weights:
\begin{equation}
\begin{aligned}
(w_\sigma,w_\pi,w_\alpha,w_h,w_\eta)
=
(0.45,0.25,0.15,0.10,0.30).
\end{aligned}
\label{eq:repair-weights}
\end{equation}
The positive terms reward counterfactual severance, root-like structural role, amplification effects, and confident descriptions, while the hazard term discourages assigning high repair value to propagated symptoms whose upstream causes remain unresolved. The final bucket is:
\begin{equation}
\rho^\sharp(r) =
\begin{cases}
\textsc{high}, & s(r)\ge 0.66,\\
\textsc{medium}, & 0.33 \le s(r) < 0.66,\\
\textsc{low}, & s(r) < 0.33.
\end{cases}
\label{eq:bucket}
\end{equation}
Two hard overrides are applied: if $r\notin\mathcal{C}_G(f^\star)$, then $\rho^\sharp(r)$ is forced to \textsc{low}; and if $\eta(r)>0$, then $\rho^\sharp(r)$ is capped at \textsc{medium}. All terms in Eq.~\eqref{eq:score} except $h(r)$ are mechanically computable from $G$, making discrepancies between the LLM-assigned bucket and the structural bucket localizable.

\subsection{Sub-Agent Delegation}
\label{sec:impl-subagent}

The top-level agent may invoke \texttt{spawn\_subagent} to delegate a narrow diagnostic question without mutating the parent state. Each sub-agent is launched with a restricted tool set
$\mathcal{U}_{\mathrm{sub}}=\{\texttt{read\_trace},\,\texttt{inspect\_state},\,\texttt{validate\_graph}\}$,
where \texttt{validate\_graph} is wrapped with \texttt{run\_auto\_repair=false} to prevent implicit state changes. The sub-agent operates on a snapshot of the parent state, with write operations intercepted so that its analysis remains effectively read-only. It returns a final JSON report containing \textit{finding}, \textit{evidence}, and \textit{suggested\_patch}. The parent agent then decides whether to apply the suggested patch through \texttt{apply\_critic\_patch}, separating diagnostic consultation from state mutation.

\subsection{Output Bundle}
\label{sec:impl-output}

Each trace produces a self-contained output bundle containing the final graph $G^\star$ as JSON (\texttt{diagnosis.json}), the full agentic step sequence (\texttt{agentic\_steps.jsonl}), per-call token and latency logs (\texttt{tool\_trace.jsonl}), aggregate metadata (\texttt{summary.json}), the console log (\texttt{run.log}), and graph visualizations in DOT, HTML, and SVG when Graphviz is available. Token accounting is handled by a \texttt{TokenAccountant}, which labels each LLM call by phase and tool invocation, enabling fine-grained cost analysis over the trajectory. These bundles serve as the unit of downstream evaluation: benchmark-level statistics and metrics can be computed by traversing the \texttt{diagnosis.json} files across a full run.

\section{CEG-Bench Dataset Analysis}
\label{app:bench-details}

This appendix provides a detailed analysis of \textbf{CEG-Bench}, the 100-trace benchmark introduced in \S\ref{sec:benchmark}. CEG-Bench is designed to evaluate causal trace diagnosis beyond final-task success, guided by the principle that a diagnostic system should identify outcome-level failures $V_F$, recover causally relevant process-level errors $V_R$ with event grounding $\varepsilon(r)$, reconstruct the typed causal propagation structure linking errors to failures, and distinguish surface anomalies $V_A$ from genuine causal contributors. The benchmark is constructed by sampling failed agentic traces, annotating them through the \textbf{Adversarial Agentic Adjudication Protocol} (AAAP), and retaining only cases with dual-adversary consensus.

During the AAAP construction process, the asymmetric four-role design is intended to reduce systematic annotation bias. A single drafter has no external mechanism for correcting its own misattributions, while a drafter paired with only one reviewer may still inherit shared model priors. AAAP therefore separates constructive graph proposal from adversarial certification: the drafter and reviewer propose and refine the CEG, while two adversaries from distinct model families independently cross-examine the result. A trace is retained only when both adversaries certify the graph, making the final annotation an intersection of two independent diagnostic perspectives. This conservative protocol rejects ambiguous cases rather than resolving them by tie-breaking, yielding the final consensus set of 100 traces.

As noted above, AAAP is still a fully LLM-based annotation protocol and therefore has inherent limitations. Although the initial pool contains 150 failed agentic traces following TraceSIR, some candidate traces are judged by AAAP to contain no clear causal error. This mainly arises from a difference in annotation setting: TraceSIR can leverage explicit failure signals such as wrong final answers or evaluator feedback when identifying errors, whereas CEG-Bench does not expose such external outcome feedback to the annotators. We make this choice to preserve the generality and fairness of CEG construction, so that the annotation process relies only on the trace itself and can be directly reproduced or extended by future users. Under this setting, even human annotators may judge some traces as acceptable if the failure is not evident from the trace alone. We therefore defer to the final AAAP consensus for inclusion decisions. Despite these limitations, we believe CEG-Bench provides a valuable reference benchmark, since its retained annotations are produced and certified through agreement between two strong agentic frameworks powered by independent frontier LLM families.

Under this construction protocol, the resulting benchmark exhibits diverse causal structures across events, errors, failures, anomalies, and typed relations. The 100 retained traces collectively contain 3{,}488 events, 391 errors, 125 failures, 253 anomalies, and 4{,}540 typed edges, including 1{,}152 non-temporal edges. The annotations cover all eight mechanisms in $\mathrm{Mech}$; the four most frequent are \textit{planning} (19.7\%), \textit{evidence integration} (18.4\%), \textit{execution} (16.9\%), and \textit{self-evaluation} (15.6\%), together accounting for 70.6\% of all errors. The role distribution is skewed toward \textsc{propagated} errors (39.6\%), followed by \textsc{root} errors (33.2\%) and \textsc{amplification} errors (27.1\%), suggesting that failures in long agentic traces often arise through cascading downstream consequences of a smaller set of root causes. Repair-value buckets are also highly imbalanced: only 3.3\% of errors receive a \textsc{high} value under $\rho^\sharp$, reflecting the strict bottleneck requirement in the counterfactual repair criterion. Detailed benchmark distributions are reported below.

\subsection{Per-Source Size Distributions}
\label{app:bench-sizes}

Table~\ref{tab:app-bench-sizes} reports the mean per-trace counts of events, errors, failures, anomalies, and edges for each source benchmark. The three sources exhibit substantially different trace profiles. Tau2Bench consists of relatively short customer-service dialogues, with a median of 14 events, where failures often trace to a small number of dominant errors. SWE-bench contains longer edit--test--debug trajectories, with a median of 35 events and a maximum of 250 events, where errors may accumulate across many tool calls. BrowseComp is the densest source in terms of errors per trace, with a mean of 6.2 errors compared with 2.0 for Tau2Bench, reflecting the multi-step evidence gathering and reasoning demands of deep-research tasks. Overall, the 50$\times$ length ratio between the longest SWE-bench trace and the shortest Tau2Bench trace stress-tests both trace normalization and the agentic inference budget $K$ from \S\ref{sec:impl-runtime}.

\begin{table}[t]
\small
\centering
\setlength{\tabcolsep}{3.5pt}
\renewcommand{\arraystretch}{1.12}
\begin{tabularx}{\columnwidth}{@{}l>{\centering\arraybackslash}X>{\centering\arraybackslash}X>{\centering\arraybackslash}X>{\centering\arraybackslash}X@{}}
\toprule
\rowcolor{gray!15}
\textbf{Component}
& \multicolumn{1}{c}{\textbf{Tau2}}
& \multicolumn{1}{c}{\textbf{SWE}}
& \multicolumn{1}{c}{\textbf{Browse}}
& \multicolumn{1}{c}{\textbf{All}} \\
\rowcolor{gray!15}
& \multicolumn{1}{c}{$n=36$}
& \multicolumn{1}{c}{$n=35$}
& \multicolumn{1}{c}{$n=29$}
& \multicolumn{1}{c}{$n=100$} \\
\midrule
Events    & 13.3 & 47.8 & 46.1 & 34.9 \\
Errors    & 2.0  & 3.9  & 6.2  & 3.9  \\
Failures  & 0.9  & 1.4  & 1.5  & 1.2  \\
Anomalies & 2.3  & 2.5  & 2.8  & 2.5  \\
Edges     & 18.0 & 57.7 & 64.6 & 45.4 \\
\bottomrule
\end{tabularx}
\caption{Mean per-trace component counts in CEG-Bench by source benchmark. Tau2, SWE, and Browse denote Tau2Bench, SWE-bench, and BrowseComp.}
\label{tab:app-bench-sizes}
\end{table}

\subsection{Vocabulary Distributions}
\label{app:bench-vocab}

\begin{table}[t]
\small
\centering
\setlength{\tabcolsep}{4pt}
\renewcommand{\arraystretch}{1.08}
\begin{tabularx}{\columnwidth}{@{}Xrr@{}}
\toprule
\textbf{Label} & \textbf{Count} & \textbf{\%} \\
\midrule
\rowcolor{gray!15}
\multicolumn{3}{@{}l}{\textbf{Mechanism $\mu$} \;(\textit{errors}, $n=391$)} \\
planning             & 77  & 19.7 \\
evidence integration & 72  & 18.4 \\
execution            & 66  & 16.9 \\
self-evaluation      & 61  & 15.6 \\
control              & 53  & 13.6 \\
omission             & 33  & 8.4  \\
representation       & 25  & 6.4  \\
environment          & 4   & 1.0  \\
\midrule
\rowcolor{gray!15}
\multicolumn{3}{@{}l}{\textbf{Structural role $\rho$} \;(\textit{errors}, $n=391$)} \\
propagated           & 155 & 39.6 \\
root                 & 130 & 33.2 \\
amplification        & 106 & 27.1 \\
\midrule
\rowcolor{gray!15}
\multicolumn{3}{@{}l}{\textbf{Repair value $\rho^\sharp$} \;(\textit{errors}, $n=391$)} \\
low                  & 244 & 62.4 \\
medium               & 134 & 34.3 \\
high                 & 13  & 3.3  \\
\midrule
\rowcolor{gray!15}
\multicolumn{3}{@{}l}{\textbf{Failure type $\theta$} \;(\textit{failures}, $n=125$)} \\
completion           & 53  & 42.4 \\
correctness          & 43  & 34.4 \\
constraint           & 17  & 13.6 \\
efficiency           & 6   & 4.8  \\
safety               & 3   & 2.4  \\
other                & 3   & 2.4  \\
\bottomrule
\end{tabularx}
\caption{Marginal distributions over the controlled vocabularies in CEG-Bench. Percentages for $\mu$, $\rho$, and $\rho^\sharp$ are computed over the 391 annotated errors; percentages for $\theta$ are computed over the 125 annotated failures.}
\label{tab:app-bench-vocab}
\end{table}

Table~\ref{tab:app-bench-vocab} summarizes the marginal distributions over the controlled vocabularies defined in \S\ref{sec:method-ceg}: error mechanism $\mu$, structural role $\rho$, failure type $\theta$, and bucketed counterfactual repair value $\rho^\sharp$. These distributions characterize the diagnostic diversity of CEG-Bench across process-level error mechanisms, causal roles, outcome-level failures, and repair priorities.
Specifically, all eight mechanisms are represented in the benchmark. The most frequent mechanisms are \textit{planning} (19.7\%), \textit{evidence integration} (18.4\%), \textit{execution} (16.9\%), and \textit{self-evaluation} (15.6\%), which together cover 70.6\% of all annotated errors. Failure types are dominated by \textit{completion} (42.4\%) and \textit{correctness} (34.4\%), with \textit{constraint} failures contributing another 13.6\%; these three categories account for 90.4\% of all failures. The role distribution is skewed toward \textsc{propagated} errors, reflecting the cascading nature of long agentic traces: many observed errors are downstream consequences of a smaller set of root causes, which the CEG structural-role typology is designed to expose.

\subsection{Edge Structure of the Causal Subgraph}
\label{app:bench-edges}

The full edge set $E$ contains 3{,}388 temporal \textsc{event-next} edges, forming one event chain per trace, and 1{,}152 non-temporal edges. The non-temporal edges consist of 403 \textsc{attached-to} grounding edges, 422 \textsc{contributes-to} error-to-failure edges, 266 \textsc{causes} edges, and 61 \textsc{amplifies} edges. Excluding temporal and grounding edges, the causal subgraph $G_C$ contains 749 causal edges, averaging 7.5 edges per trace. The relative sparsity of \textsc{amplifies} edges compared with \textsc{causes} edges, at roughly $1:4.4$, indicates that amplification is a specialized relation rather than a generic propagation link. Most causal structure in CEG-Bench therefore flows through error propagation chains that terminate in \textsc{contributes-to} edges to failures, while \textsc{amplifies} is reserved for cases where one error meaningfully magnifies the impact of another.

\subsection{Repair-Value Distribution}
\label{app:bench-repair}

\begin{table}[t]
\footnotesize
\centering
\setlength{\tabcolsep}{3.5pt}
\renewcommand{\arraystretch}{1.10}
\begin{tabularx}{\columnwidth}{@{}Xrrrr@{}}
\toprule
\rowcolor{gray!15}
\textbf{Mechanism}
& \textbf{\textsc{high}}
& \textbf{\textsc{med.}}
& \textbf{\textsc{low}}
& \textbf{Total} \\
\midrule
planning             & 0 & 40 & 37 & 77 \\
evidence integration & 3 & 19 & 50 & 72 \\
execution            & 3 & 18 & 45 & 66 \\
self-evaluation      & 0 & 14 & 47 & 61 \\
control              & 1 & 15 & 37 & 53 \\
omission             & 2 & 11 & 20 & 33 \\
representation       & 0 & 17 & 8  & 25 \\
environment          & 4 & 0  & 0  & 4  \\
\midrule
\rowcolor{gray!15}
\textbf{Total}       & \textbf{13} & \textbf{134} & \textbf{244} & \textbf{391} \\
\bottomrule
\end{tabularx}
\caption{Repair-value buckets cross-tabulated by error mechanism. Counts are computed over the 391 annotated errors in CEG-Bench.}
\label{tab:app-repair-xtab}
\end{table}

The bucketed counterfactual repair values $\rho^\sharp$ from \S\ref{sec:impl-repair-value} are highly long-tailed: 244 errors (62.4\%) fall in \textsc{low}, 134 errors (34.3\%) in \textsc{medium}, and only 13 errors (3.3\%) in \textsc{high}. This skew follows from the repair-value rubric. To receive a \textsc{high} value, an error must act as a strong counterfactual bottleneck for the primary failure $f^\star$, reflected by a large severance score $\sigma(r,f^\star)$ in Eq.~\eqref{eq:severance}, and must also avoid the hard overrides in the bucketing rule of Eq.~\eqref{eq:bucket}.
Table~\ref{tab:app-repair-xtab} cross-tabulates repair-value buckets against error mechanisms. The 13 \textsc{high}-valued errors are concentrated in \textit{environment}, \textit{evidence integration}, \textit{execution}, and \textit{omission}. In particular, all four \textit{environment} errors receive \textsc{high} values, while the remaining high-value cases typically correspond to situations where a single missing, incorrect, or unavailable action blocks task completion under the annotated causal graph. Thus, CEG-Bench provides not only diagnostic structure but also a built-in repair-prioritization signal for downstream debugging methods.

\section{CEG Evaluation Protocols}
\label{app:eval-spec}

This appendix specifies the two evaluation protocols introduced in \S\ref{sec:bench-eval}. We release both as reference implementations.

\subsection{Semantically Relaxed CEG Evaluation}
\label{app:eval-relaxed}

The semantically relaxed evaluator first aligns predicted and gold nodes across graphs using bipartite matching. It then translates predicted identifiers into the gold identifier space, computes precision, recall, and F1 for each scoring dimension, and aggregates the resulting dimension scores into a tier-weighted composite.

\paragraph{Node alignment.}
For each node type $c\in\{\text{event},\text{error},\text{failure},\text{anomaly}\}$, let $V_c^\star$ denote the gold node set and $\hat{V}_c$ the predicted node set. We construct a similarity matrix
$\mathbf{M}_c\in[0,1]^{|V_c^\star|\times|\hat{V}_c|}$ and solve a maximum-weight bipartite matching problem:
\begin{equation}
\begin{aligned}
\Lambda_c^\star
=
\arg\max_{\Lambda \in \mathrm{Match}(V_c^\star,\hat{V}_c)}
\sum_{(g,p)\in\Lambda} \mathbf{M}_c[g,p].
\end{aligned}
\label{eq:app-hungarian}
\end{equation}
Only matched pairs satisfying $\mathbf{M}_c[g,p]\ge \tau_c$ are retained. The default type-specific thresholds are
$\tau_{\mathrm{event}}=\tau_{\mathrm{error}}=\tau_{\mathrm{failure}}=\tau_{\mathrm{anomaly}}=0.60$.

\paragraph{Per-node similarity.}
Node similarity is type-specific. For error nodes, we use the following empirical similarity function:
\begin{equation}
\begin{aligned}
&\mathrm{sim}(r,\hat{r})
= {}
w_e\,\mathrm{sim}_E\bigl(\varepsilon(r),\varepsilon(\hat{r})\bigr) \\
&+ w_\mu\,\mathbb{1}[\mu(r)=\mu(\hat{r})]  + w_\rho\,\mathbb{1}[\rho(r)=\rho(\hat{r})] \\
&+ w_\nu\,\mathrm{sim}_T\bigl(\nu(r),\nu(\hat{r})\bigr) + w_d\,\mathrm{sim}_T\bigl(d(r),d(\hat{r})\bigr),
\end{aligned}
\label{eq:app-err-sim}
\end{equation}
with weights
$(w_e,w_\mu,w_\rho,w_\nu,w_d)=(0.25,0.20,0.15,0.15,0.25)$. We set these weights empirically; in practice, the final scores are not sensitive to small perturbations of them. Here $\mathrm{sim}_E$ denotes cross-graph event-alignment similarity, computed using an inner Hungarian alignment over events together with an order-consistency factor based on the longest non-decreasing subsequence of matched event indices. For free-text fields, $\mathrm{sim}_T$ is computed as cosine similarity between BGE-M3 embeddings \citep{bge-m3}. Other node types use the corresponding subset of categorical, event, and textual fields.

\paragraph{Dimensional scoring.}
After node alignment, predicted identifiers are translated into gold identifiers through the matched pairs. Each scoring dimension is then evaluated using precision, recall, and F1 over either typed node multisets or translated edge tuples. The relaxed evaluator reports eight dimensions grouped into three tiers:
\begin{itemize}\setlength{\itemsep}{1pt}
    \item \textbf{Tier 1: structural dimensions} with total weight $0.60$: \texttt{error\_node}, \texttt{causal\_edge}, and \texttt{main\_chain}, each weighted by $0.20$.
    \item \textbf{Tier 2: textual and sparse dimensions} with total weight $0.30$: \texttt{failure}, \texttt{attribution}, and \texttt{anomaly}, each weighted by $0.10$.
    \item \textbf{Tier 3: utility dimensions} with total weight $0.10$: \texttt{event} and \texttt{consistency}, each weighted by $0.05$.
\end{itemize}
The \texttt{main\_chain} dimension is the only non-elementary dimension. For each graph, we extract a canonical root-to-failure path in the causal subgraph $G_C$, defined as the longest weighted path from any \textsc{root} error to the primary failure $f^\star$ along causal labels in $\mathcal{L}_C$. Chain similarity is then computed by a weighted combination of node overlap, edge overlap, and length-normalized longest common subsequence (LCS) over the matched error sequence. The \texttt{consistency} dimension checks intra-graph structural invariants, such as the reachability constraint $\mathrm{W}_3$, and penalizes violations in $\hat{G}$.

\paragraph{Composite score.}
The final relaxed score is the tier-weighted sum of dimension-level F1 scores:
\begin{equation}
\mathrm{CEG\text{-}Sim}(\hat{G},G^\star)
=
\sum_{c} w_c\,\mathrm{F1}_c
\in[0,1],
\label{eq:app-relaxed-overall}
\end{equation}
where $\sum_c w_c=1$. We use $\mathrm{CEG\text{-}Sim}$ as the headline semantically relaxed metric.

\subsection{Structurally Exact CEG Evaluation}
\label{app:eval-exact}

The structurally exact evaluator is rule-based, LLM-free, and performs no fuzzy matching. It asks whether a prediction $\hat{G}$ reconstructs the consensus graph $G^\star$ as a structured object, rather than merely matching it semantically.

\paragraph{Cross-graph identifier resolution.}
Gold and predicted CEGs use independent local identifier spaces for errors and failures, but both refer to the same shared event identifiers $V_E$ produced by the deterministic trace parser in \S\ref{sec:impl-trace}. We therefore resolve cross-graph identifiers without text comparison: errors are matched by the exact tuple key $(\varepsilon(r),\mu(r))$, combining event grounding and controlled mechanism, while failures are matched by the type bucket $\theta(f)$. The resulting maps $\phi_R:\hat{V}_R\to V_R^\star$ and $\phi_F:\hat{V}_F\to V_F^\star$ translate predicted nodes and edges into the gold identifier space.

\paragraph{Dimensional scoring.}
The evaluator scores seven dimensions, each using precision, recall, and F1 over finite sets or multisets of typed tuples:
\begin{itemize}\setlength{\itemsep}{1pt}
    \item \textbf{event}: identifier-set match over $V_E$.
    \item \textbf{event\_next}: edge-set match over $\{(u,v):(u,\textsc{event-next},v)\in E\}$.
    \item \textbf{failure}: multiset match over failure buckets $\theta(f)$, including count agreement.
    \item \textbf{error}: match over $(\varepsilon(r),\mu(r))$ tuples, with an additional $\rho(r)$ agreement check on matched pairs.
    \item \textbf{attached\_to}: translated \textsc{attached-to} edge-set match using $\phi_R$.
    \item \textbf{causes}: translated match over the union of \textsc{causes} and \textsc{amplifies} edges using $\phi_R$.
    \item \textbf{contributes\_to}: translated error-to-failure edge-set match using $\phi_R$ and $\phi_F$.
\end{itemize}
Predicted edges incident to unmatched errors or failures are counted as false positives. This propagates node-resolution errors into downstream edge dimensions without introducing additional cascading penalties.

\paragraph{Composite score.}
The composite score is the unweighted mean of five model-discriminative dimensions. Let $
\mathcal{D}^\star =
\{$\texttt{failure}, \texttt{error}, \texttt{attached\_to}, \texttt{causes}, \texttt{contributes\_to}$\}$.
Then,
\begin{equation}
\mathrm{CEG\text{-}Exact}(\hat{G},G^\star)
=
\frac{1}{|\mathcal{D}^\star|}
\sum_{c\in\mathcal{D}^\star}\mathrm{F1}_c.
\label{eq:app-exact-overall}
\end{equation}
The two parser-driven dimensions, \texttt{event} and \texttt{event\_next}, are reported but excluded from the composite because they are determined by the shared trace parser rather than by model capability.

\section{Human Gold Validation of CEG-Bench}
\label{app:human-gold}

The annotations in \textbf{CEG-Bench} are produced by agentic systems through AAAP rather than by exhaustive human annotation. A natural concern is therefore whether these automatically constructed CEGs are reliable. To provide an independent audit, we construct \textbf{CEG-Bench-Gold}, an expert-curated human reference set over the same 100 traces. This appendix describes the construction of CEG-Bench-Gold, measures the agreement between AAAP and human annotations, and re-evaluates all systems against the human reference under the same SRE and SEE criteria used in the main paper.

\paragraph{Construction of CEG-Bench-Gold.}
We invited eight domain experts, each with graduate-level training in computer science, to annotate and verify the 100 traces in CEG-Bench. Given the raw trace and the CEG schema in \S\ref{sec:method-ceg}, experts marked events, failures, anomalies, causally relevant errors with their mechanisms and roles, and the causal edges among errors and failures. Each trace was reviewed by multiple experts, and disagreements were resolved through discussion into a consensus human graph. Constructing CEGs is difficult even for experts because agentic traces are long, heterogeneous, and often causally ambiguous. We therefore use CEG-Bench-Gold as a carefully curated audit set for validating AAAP, rather than as a scalable replacement for automated annotation. We release CEG-Bench-Gold together with the AAAP-annotated CEG-Bench.

\paragraph{AAAP agreement with the human gold.}
We quantify the reliability of AAAP by scoring the AAAP annotations against CEG-Bench-Gold, treating the human graph as the reference and the AAAP graph as the prediction. The same SRE and SEE metrics from \S\ref{sec:bench-eval} are used. As shown in Figure~\ref{fig:gold-similarity} and Table~\ref{tab:app-human-agreement}, AAAP shows strong agreement with expert judgments. It achieves an overall $\mathrm{CEG\text{-}Sim}$ of $71.5$, with high SRE agreement on anomalies ($88.1$), errors ($76.3$), and failures ($76.3$). Agreement is lower on the strictest causal-wiring dimensions, such as SEE \texttt{causes} ($20.1$) and \texttt{contributes\_to} ($36.2$), which require exact reconstruction of fine-grained graph edges and are difficult even under careful human adjudication. Notably, SEE agreement increases with trace complexity, from $43.0$ on small traces to $45.8$ on medium traces and $52.4$ on large traces, suggesting that AAAP remains reliable on the long, causally rich traces for which structured diagnosis is most needed.

\begin{table}[t]
\centering
\small
\setlength{\tabcolsep}{4pt}
\renewcommand{\arraystretch}{1.1}
\begin{tabularx}{\columnwidth}{@{}l l >{\centering\arraybackslash}X@{}}
\toprule
\rowcolor{gray!15}
\textbf{Criterion} & \textbf{Dimension} & \textbf{Agreement} \\
\midrule
\multirow{8}{*}{SRE}
 & \cellcolor{blue!8}\textbf{CEG-Sim (overall)} & \cellcolor{blue!8}\textbf{71.5} \\
 & anomaly            & 88.1 \\
 & error              & 76.3 \\
 & failure            & 76.3 \\
 & main\_chain        & 61.9 \\
 & attribution        & 58.8 \\
 & causal\_edge       & 58.0 \\
 & event              & 99.0 \\
\midrule
\multirow{6}{*}{SEE}
 & \cellcolor{blue!8}\textbf{overall}          & \cellcolor{blue!8}\textbf{47.1} \\
 & failure            & 76.6 \\
 & error              & 52.4 \\
 & attached\_to       & 50.2 \\
 & contributes\_to    & 36.2 \\
 & causes             & 20.1 \\
\bottomrule
\end{tabularx}
\caption{Agreement between the AAAP annotations in CEG-Bench and the expert-curated CEG-Bench-Gold, measured with SRE and SEE over all 100 traces (\%). SEE agreement by complexity tercile is $43.0/45.8/52.4$ for small/medium/large traces.}
\label{tab:app-human-agreement}
\end{table}

\paragraph{System scores against the human gold.}
To check whether our main conclusions depend on the AAAP reference, we additionally evaluate every system against CEG-Bench-Gold using the same SRE and SEE criteria. Tables~\ref{tab:app-gold-sre} and~\ref{tab:app-gold-see} mirror the two main result tables, but replace the AAAP annotations with the human reference. The within-family ranking is preserved: \textbf{CEG-Agent obtains the best composite SRE and SEE within both backbone families}. This confirms that the observed advantage of CEG-Agent is not an artifact of evaluating against agent-generated annotations.

As in the main results, the only dimension where a baseline remains competitive is SEE \texttt{failure}, where systems that emit a small number of failure nodes can be mechanically favored on some traces (\S\ref{app:exp-empty}). Absolute scores are generally lower than those obtained against the AAAP reference, which is expected because CEG-Bench-Gold is an independent and stricter reference. Importantly, this shift lowers all systems together without changing the main ordering, supporting the robustness of our conclusions.

\begin{table*}[t]
\centering
\small
\setlength{\tabcolsep}{3.0pt}
\renewcommand{\arraystretch}{1.12}
\begin{tabularx}{\textwidth}{@{}
>{\raggedright\arraybackslash}p{2.35cm}
>{\raggedright\arraybackslash}p{3.25cm}
c
YYYYYYYY}
\toprule
\rowcolor{gray!12}
\textbf{Backbone} & \textbf{System}
& \cellcolor{blue!14}\textbf{SRE}
& \multicolumn{3}{c}{\textbf{Structural}}
& \multicolumn{3}{c}{\textbf{Textual}}
& \multicolumn{2}{c}{\textbf{Utility}} \\
\rowcolor{gray!12}
& & & \texttt{error} & \texttt{edge} & \texttt{chain}
& \texttt{fail.} & \texttt{attr.} & \texttt{anom.} & \texttt{event} & \texttt{cons.} \\
\cmidrule(lr){3-3}\cmidrule(lr){4-6}\cmidrule(lr){7-9}\cmidrule(l){10-11}
Claude-Opus-4.7 & Claude Code (raw)
& \totalcell{38.1} & 36.3 & 26.5 & 24.8 & \textbf{49.1} & 24.4 & 43.4 & 61.4 & \textbf{100.0} \\
Claude-Opus-4.7 & Claude Code (+Process)
& \totalcell{38.1} & 35.6 & 26.0 & 23.6 & 40.5 & 23.8 & 46.5 & \textbf{99.0} & \textbf{100.0} \\
\rowcolor{blue!3}
Claude-Opus-4.7 & \textbf{CEG-Claude} (ours)
& \besttotal{45.4} & \textbf{43.2} & \textbf{30.2} & \textbf{37.1} & 45.0 & \textbf{30.6} & \textbf{57.6} & \textbf{99.0} & 99.9 \\
\addlinespace[2pt]
\midrule
GPT-5.5 & Codex (raw)
& \totalcell{27.7} & 21.4 & 22.2 & 15.1 & 38.3 & 16.5 & 6.9 & 99.0 & 100.0 \\
GPT-5.5 & Codex (+Process)
& \totalcell{26.6} & 21.7 & 23.5 & 15.1 & 26.6 & 12.4 & 6.9 & 99.0 & 100.0 \\
\rowcolor{blue!3}
GPT-5.5 & \textbf{CEG-Codex} (ours)
& \besttotal{40.8} & \textbf{38.1} & \textbf{28.0} & \textbf{29.3} & \textbf{44.6} & \textbf{24.9} & \textbf{48.8} & 99.0 & 100.0 \\
\bottomrule
\end{tabularx}
\caption{Semantically relaxed evaluation (\textbf{SRE}) against the expert-curated \textbf{CEG-Bench-Gold}, using the same layout as Table~\ref{tab:exp-sre}. Best score within each backbone block is in \textbf{bold}. Scores are lower than those against the AAAP reference, but the within-family ranking is preserved.}
\label{tab:app-gold-sre}
\end{table*}

\begin{table*}[t]
\centering
\small
\setlength{\tabcolsep}{4.0pt}
\renewcommand{\arraystretch}{1.12}
\begin{tabularx}{\textwidth}{@{}
>{\raggedright\arraybackslash}p{2.35cm}
>{\raggedright\arraybackslash}p{3.25cm}
c
YYYYY}
\toprule
\rowcolor{gray!12}
\textbf{Backbone} & \textbf{System}
& \cellcolor{blue!14}\textbf{SEE}
& \multicolumn{5}{c}{\textbf{Dimensions}} \\
\rowcolor{gray!12}
& & & \texttt{failure} & \texttt{error} & \texttt{attach} & \texttt{cause} & \texttt{contrib.} \\
\cmidrule(lr){3-3}\cmidrule(l){4-8}
Claude-Opus-4.7 & Claude Code (raw)
& \totalcell{15.5} & \textbf{60.1} & 8.5 & 6.1 & 0.5 & 2.2 \\
Claude-Opus-4.7 & Claude Code (+Process)
& \totalcell{19.5} & 49.0 & 18.4 & 17.4 & 1.3 & 11.3 \\
\rowcolor{blue!3}
Claude-Opus-4.7 & \textbf{CEG-Claude} (ours)
& \besttotal{22.6} & 57.9 & \textbf{22.4} & \textbf{18.3} & \textbf{2.0} & \textbf{12.4} \\
\addlinespace[2pt]
\midrule
GPT-5.5 & Codex (raw)
& \totalcell{16.3} & 55.3 & 8.9 & 8.7 & 0.0 & 8.6 \\
GPT-5.5 & Codex (+Process)
& \totalcell{13.3} & 39.5 & 9.2 & 8.4 & 1.7 & 7.5 \\
\rowcolor{blue!3}
GPT-5.5 & \textbf{CEG-Codex} (ours)
& \besttotal{22.4} & \textbf{59.5} & \textbf{18.5} & \textbf{17.6} & \textbf{2.3} & \textbf{14.1} \\
\bottomrule
\end{tabularx}
\caption{Structurally exact evaluation (\textbf{SEE}) against the expert-curated \textbf{CEG-Bench-Gold}, using the same layout as Table~\ref{tab:exp-see}. Best score within each backbone block is in \textbf{bold}.}
\label{tab:app-gold-see}
\end{table*}

\paragraph{Why the main paper uses AAAP as the primary reference.}
The human-gold evaluations above confirm that CEG-Agent's advantage holds under an independent expert reference. We nevertheless use the AAAP-constructed CEG-Bench as the primary benchmark in the main paper for two reasons. First, scalable automated annotation is a central contribution of this work: high-quality human annotation of long causal traces is expensive, slow, and rarely available in real deployments, whereas AAAP is fully automated and, as shown above, closely agrees with expert judgments. Second, using a single AAAP reference throughout the paper keeps the benchmark analyses, ablations, cross-backbone comparisons, and closed-loop repair studies directly comparable. We therefore treat CEG-Bench-Gold as an independent audit of AAAP reliability, while reporting the main experiments against the AAAP annotations.

\section{Experimental Analyses}
\label{app:exp-details}

This appendix section expands \S\ref{sec:exp} with the full set of experimental analyses that are summarized in the main paper. It contains a detailed experimental setup (\S\ref{app:exp-setup-details}), per-source results (\S\ref{app:exp-source}), robustness to trace complexity (\S\ref{app:exp-complexity}), full dimension-level diagnosis (\S\ref{app:exp-dim}), abstention behaviour on gold-empty traces (\S\ref{app:exp-empty}), graph-size calibration (\S\ref{app:exp-calibration}), per-mechanism / per-role recall (\S\ref{app:exp-mech-role}), failure-type coverage (\S\ref{app:exp-failuretype}), repair-value calibration (\S\ref{app:exp-repair}), inter-system agreement on irreducibly hard traces (\S\ref{app:exp-agreement}), a per-backbone diagnostic capability profile (\S\ref{app:exp-backbone-detail}), the cost and tool-use profile (\S\ref{app:exp-cost}), and a qualitative case study (\S\ref{app:exp-case}).

\subsection{Detailed Experimental Setup}
\label{app:exp-setup-details}

\paragraph{Systems.}
We compare CEG-Agent against strong agentic baselines on identical input traces. For each backbone family, we evaluate three systems:
\begin{itemize}\setlength{\itemsep}{1pt}
    \item \textbf{Agentic baseline (raw).} An off-the-shelf coding agent, either Claude Code or Codex, receives the raw trace JSON together with the CEG schema and is asked to produce a single \texttt{diagnosis.json}. It must perform event segmentation and causal diagnosis end-to-end.
    \item \textbf{Agentic baseline (+Process).} The same agent receives traces that have already been segmented into events $V_E$ and connected by \textsc{event-next} edges using our deterministic parser (\S\ref{sec:impl-trace}). This setting isolates causal diagnosis from trace parsing and provides a stronger baseline for comparison.
    \item \textbf{CEG-Agent (ours).} The full tool-augmented framework described in \S\ref{sec:method}, instantiated with the same backbone LLM as the corresponding baseline.
\end{itemize}
We use Claude-Opus-4.7 for the Claude Code / CEG-Claude pair and GPT-5.5 for the Codex / CEG-Codex pair. All systems are evaluated on the same 100 traces and against the same consensus CEG annotations.

\paragraph{Evaluation.}
We report the two complementary criteria defined in \S\ref{sec:bench-eval}: the semantically relaxed evaluator $\mathrm{CEG\text{-}Sim}$ (\textbf{SRE}) and the structurally exact evaluator $\mathrm{CEG\text{-}Exact}$ (\textbf{SEE}), together with per-dimension breakdowns. Higher scores are better, and all values are reported as percentages.

\paragraph{Hyperparameters and scoring.}
CEG-Agent uses inference budget $K=24$, critic budget $C=2$, temperature $=1$, and repair-value estimation enabled. Baselines are run with their respective coding agents' default CLI configurations. 

\subsection{Per-Source Performance}
\label{app:exp-source}

\begin{table*}[t]
\centering
\small
\setlength{\tabcolsep}{4.0pt}
\renewcommand{\arraystretch}{1.12}
\begin{tabularx}{\textwidth}{@{}
>{\raggedright\arraybackslash}p{2.35cm}
>{\centering\arraybackslash}p{2.0cm}
>{\raggedright\arraybackslash}p{3.35cm}
YYY
@{}}
\toprule
\rowcolor{gray!12}
\textbf{Backbone} & \textbf{Evaluation} & \textbf{System}
& \textbf{Tau2Bench} & \textbf{SWE-bench} & \textbf{BrowseComp} \\
\rowcolor{gray!12}
& & & $n=36$ & $n=35$ & $n=29$ \\
\midrule
Claude-Opus-4.7 & SRE & Claude Code (raw)
& 59.8 & 37.3 & 38.7 \\
Claude-Opus-4.7 & SRE & Claude Code (+Process)
& \textbf{67.2} & 38.3 & 41.5 \\
\rowcolor{blue!3}
Claude-Opus-4.7 & SRE & \textbf{CEG-Claude} (ours)
& 60.0 & \textbf{47.7} & \textbf{45.9} \\
\addlinespace[2pt]
Claude-Opus-4.7 & SEE & Claude Code (raw)
& 16.8 & 11.1 & 13.8 \\
Claude-Opus-4.7 & SEE & Claude Code (+Process)
& \textbf{24.9} & 13.5 & 19.3 \\
\rowcolor{blue!3}
Claude-Opus-4.7 & SEE & \textbf{CEG-Claude} (ours)
& 20.7 & \textbf{19.7} & \textbf{21.0} \\
\midrule
GPT-5.5 & SRE & Codex (raw)
& 41.5 & 24.6 & 29.3 \\
GPT-5.5 & SRE & Codex (+Process)
& 34.4 & 33.1 & 30.7 \\
\rowcolor{blue!3}
GPT-5.5 & SRE & \textbf{CEG-Codex} (ours)
& \textbf{56.2} & \textbf{41.2} & \textbf{40.7} \\
\addlinespace[2pt]
GPT-5.5 & SEE & Codex (raw)
& 12.8 & 7.8  & \textbf{15.1} \\
GPT-5.5 & SEE & Codex (+Process)
& 2.1  & 15.1 & \textbf{15.1} \\
\rowcolor{blue!3}
GPT-5.5 & SEE & \textbf{CEG-Codex} (ours)
& \textbf{17.3} & \textbf{17.1} & 16.1 \\
\bottomrule
\end{tabularx}
\caption{Per-source composite scores on CEG-Bench. Best score within each backbone--evaluator block is in \textbf{bold}.}
\label{tab:app-exp-source}
\end{table*}

Table~\ref{tab:app-exp-source} shows that the benefit of CEG-Agent depends on source complexity. Under Claude-Opus-4.7, Claude Code (+Process) is strongest on the short dialogue-style Tau2Bench traces, where the task often reduces to attaching one or two errors to a small event scaffold. On the longer SWE-bench traces and the denser BrowseComp traces, the pattern reverses: CEG-Claude outperforms the +Process baseline by $+9.4$ and $+4.4$ points in SRE, and by $+6.2$ and $+1.7$ points in SEE, respectively. The Tau2Bench regression is largely explained by the gold-empty abstention pattern analyzed in \S\ref{app:exp-empty}.

Under GPT-5.5, the improvement is more uniform. CEG-Codex achieves the best SRE score on all three sources and the best SEE score on two of three sources, including a $+21.8$-point SRE gain over Codex (+Process) on Tau2Bench. The weak Tau2Bench SEE score of Codex (+Process) reflects a tendency to hallucinate errors when given an event scaffold without sufficient diagnostic grounding; CEG-Agent mitigates this failure mode through structured graph construction and validation (\S\ref{app:exp-calibration}).

\subsection{Robustness to Trace Complexity}
\label{app:exp-complexity}

\begin{table*}[t]
\centering
\small
\setlength{\tabcolsep}{4.0pt}
\renewcommand{\arraystretch}{1.12}
\begin{tabularx}{\textwidth}{@{}
>{\raggedright\arraybackslash}p{2.35cm}
>{\raggedright\arraybackslash}p{3.25cm}
YYY
@{}}
\toprule
\rowcolor{gray!12}
\textbf{Backbone} & \textbf{System}
& \textbf{Small}
& \textbf{Medium}
& \textbf{Large} \\
\rowcolor{gray!12}
& 
& $\bar{|V_E|}\approx 9$
& $\bar{|V_E|}\approx 23$
& $\bar{|V_E|}\approx 72$ \\
\midrule
Claude-Opus-4.7
& Claude Code (raw)
& 16.6 & 10.9 & 14.2 \\
Claude-Opus-4.7
& Claude Code (+Process)
& \textbf{23.0} & 14.9 & 20.0 \\
\rowcolor{blue!3}
Claude-Opus-4.7
& \textbf{CEG-Claude} (ours)
& 22.9 & \textbf{17.0} & \textbf{21.4} \\
\midrule
GPT-5.5
& Codex (raw)
& 10.7 & 9.3 & 15.0 \\
GPT-5.5
& Codex (+Process)
& 9.1 & 3.9 & \textbf{18.0} \\
\rowcolor{blue!3}
GPT-5.5
& \textbf{CEG-Codex} (ours)
& \textbf{17.7} & \textbf{15.0} & 17.8 \\
\bottomrule
\end{tabularx}
\caption{SEE composite by trace-complexity tercile. Best score within each backbone block is in \textbf{bold}.}
\label{tab:app-exp-complexity}
\end{table*}

Table~\ref{tab:app-exp-complexity} groups CEG-Bench traces into terciles by gold event count and reports SEE composite scores. Under Claude-Opus-4.7, the +Process baseline is strongest on small traces, but degrades on medium traces and remains below CEG-Claude on medium and large traces. CEG-Claude shows a flatter curve, suggesting that incremental graph construction and schema-guided repair help stabilize diagnosis as traces grow longer.
The effect is more pronounced under GPT-5.5. Codex (+Process) collapses on medium-length traces, dropping to 3.9 SEE, while CEG-Codex maintains 15.0, an $+11.1$ point gain in the regime where multiple errors must be assembled into a coherent causal structure. On large traces, CEG-Codex remains competitive with the best baseline. Overall, these results indicate that CEG-Agent is most beneficial when trace diagnosis requires sustained causal graph construction rather than isolated error spotting.

\subsection{Dimension-Level Diagnosis}
\label{app:exp-dim}

Tables~\ref{tab:exp-sre} and \ref{tab:exp-see} provide the dimension-level results; here we extend the analysis to understand where the framework gains come from.

\paragraph{Causal-structure recovery is the main source of gain.}
Across both backbone families, CEG-Agent improves the dimensions most directly tied to causal error reconstruction. The largest gains appear on SRE \texttt{main\_chain}, where CEG-Claude improves over the Claude baselines by $+12.2$/$+12.8$ points and CEG-Codex improves over the Codex baselines by $+14.1$/$+12.0$ points. CEG-Agent also improves the structurally exact dimensions that require precise error grounding and causal wiring, especially SEE \texttt{error}, \texttt{attached\_to}, and \texttt{causes}. These dimensions directly measure whether a system can recover the process-level errors, ground them to trace events, and reconstruct the causal links through which they contribute to failures, which is precisely the role of CEG-Agent's validator-guided graph construction and schema projection.

\paragraph{Anomaly--error separation is a robust win.}
SRE \texttt{anomaly} improves from $44.7/46.0$ to $60.0$ under Claude and from near-zero $2.0/1.7$ to $51.3$ under GPT-5.5. This reflects the benefit of CEG-Agent's explicit anomaly--error--failure taxonomy. Baselines frequently collapse surface irregularities into process-level errors or omit anomalies altogether; the Codex baselines almost never emit anomaly nodes. In contrast, CEG-Agent uses a dedicated anomaly-detection step, allowing it to recover surface irregularities without conflating them with causally relevant errors.

\paragraph{Textual failure dimensions trade off only under Claude.}
CEG-Claude underperforms Claude Code (+Process) on the SRE textual dimensions \texttt{failure} and \texttt{attribution} (38.8 vs.\ 47.7 and 37.1 vs.\ 43.9). These dimensions rely on embedding-based textual similarity, where the gold annotations and +Process outputs are often concise, while CEG-Claude produces more detailed failure localization and explanatory descriptions. Although such detail can be useful for debugging, it may reduce similarity to compact gold references. The trade-off does not persist under structurally exact evaluation: CEG-Claude outperforms +Process on SEE \texttt{failure} by $+4.8$ points, and it disappears under GPT-5.5, where CEG-Codex leads both Codex baselines on SRE \texttt{failure} and \texttt{attribution}. This suggests a textual-style mismatch rather than a limitation of the CEG-Agent framework.

\paragraph{Where structural gaps remain.}
The main remaining gap is SEE \texttt{contributes\_to}. CEG-Claude trails Claude Code (+Process) on this dimension (9.9 vs.\ 13.2), and although CEG-Codex leads its baselines, the margin is smaller than on other structural dimensions. This occurs because CEG-Agent often emits a richer set of error nodes, but does not always attach every newly introduced error to the failure node with an explicit \textsc{contributes-to} edge. The schema projection restores reachability when needed, but exact tuple matching still rewards compact graphs with fewer missing or extra failure-link edges. Improving \textsc{contributes-to} coverage is therefore a natural target for future refinement.

\begin{table*}[t]
\centering
\small
\setlength{\tabcolsep}{3.2pt}
\renewcommand{\arraystretch}{1.12}
\begin{tabularx}{\textwidth}{@{}
>{\raggedright\arraybackslash}p{2.35cm}
>{\raggedright\arraybackslash}p{3.25cm}
YYYYYY}
\toprule
\rowcolor{gray!12}
\textbf{Backbone} & \textbf{System}
& \multicolumn{2}{c}{\textbf{SRE}}
& \multicolumn{2}{c}{\textbf{SEE}}
& \multicolumn{2}{c}{\textbf{Gold-empty}} \\
\rowcolor{gray!12}
&
& \texttt{empty}
& \texttt{non-empty}
& \texttt{empty}
& \texttt{non-empty}
& \texttt{abstain}
& \texttt{error} \\
\cmidrule(lr){3-4}
\cmidrule(lr){5-6}
\cmidrule(l){7-8}
Claude-Opus-4.7
& Claude Code (raw)
& 64.8 & 42.5 & 0.0 & 16.4 & 1/15  & 14/15 \\
Claude-Opus-4.7
& Claude Code (+Process)
& \textbf{87.1} & 43.0 & 0.0 & 22.7 & {11/15} & 4/15 \\
\rowcolor{blue!3}
Claude-Opus-4.7
& \textbf{CEG-Claude} (ours)
& 68.6 & \textbf{48.6} & 0.0 & \textbf{24.1} & 0/15 & 14/15 \\
\midrule
GPT-5.5
& Codex (raw)
& 59.9 & 27.1 & 0.0 & 13.7 & 0/15 & 15/15 \\
GPT-5.5
& Codex (+Process)
& 60.0 & 28.1 & 0.0 & 12.3 & 0/15 & 15/15 \\
\rowcolor{blue!3}
GPT-5.5
& \textbf{CEG-Codex} (ours)
& \textbf{77.1} & \textbf{41.1} & 0.0 & \textbf{19.9} & 0/15 & 7/15 \\
\bottomrule
\end{tabularx}
\caption{Performance split by gold-empty and non-empty CEGs. \texttt{abstain} counts empty predictions; \texttt{error} counts predictions with at least one error on gold-empty traces.}
\label{tab:app-exp-empty}
\end{table*}

\subsection{Gold-Empty Traces and Abstention Behavior}
\label{app:exp-empty}

A subset of $15/100$ CEG-Bench traces, disproportionately from Tau2Bench ($10/36$), are retained with \emph{empty} CEGs because the dual-adversary AAAP could not identify causal evidence from the trace alone (\S\ref{app:bench-details}). These cases primarily test abstention rather than diagnosis: systems that emit empty graphs are mechanically rewarded, while systems that propose causal errors are penalized. Table~\ref{tab:app-exp-empty} separates results on gold-empty and non-empty traces.

\paragraph{Abstention is backbone-dependent.}
Claude Code (+Process) is the only system that frequently abstains, returning an empty graph on $11/15$ gold-empty traces and reaching SRE $87.1$ on this subset. This accounts for much of its Tau2Bench advantage over CEG-Claude in Table~\ref{tab:app-exp-source}. By contrast, CEG-Claude almost always emits a diagnosis, and both Codex baselines emit errors on all gold-empty traces. 

\paragraph{CEG-Agent improves on substantive traces.}
On the $n=85$ non-empty traces, where causal diagnosis is required, both CEG-Agent variants achieve higher composite scores than their same-backbone baselines. CEG-Claude improves by $+5.6$ SRE and $+1.4$ SEE, while CEG-Codex improves by $+13.0$ SRE and $+6.2$ SEE. This indicates that the framework's advantage is concentrated on traces with substantive causal structure, and that the Claude-family Tau2Bench regression is mainly an abstention artifact rather than a diagnostic failure. A natural improvement is to add a calibrated empty-graph option to the failure-analysis stage, allowing CEG-Agent to abstain when no causal error is supported by the trace. We leave this out deliberately: for generality, we want CEG-Agent to focus on diagnosing and constructing CEGs for substantive traces, rather than tuning its behavior around a separate empty-case classification task.

\paragraph{Per-dimension breakdown on empty and non-empty subsets.}
Tables~\ref{tab:app-exp-empty-sre-dim} and \ref{tab:app-exp-empty-see-dim} provide the per-dimension breakdown. On the gold-empty subset, several SRE dimensions are mechanically saturated because the gold graph contains no causal structure; the most informative columns are therefore \texttt{error}, \texttt{failure}, and \texttt{anomaly}. Claude Code (+Process) scores highly on this subset largely because it abstains on most gold-empty traces. SEE is zero for all systems on the empty subset because there are no gold errors, failures, or causal edges to match.

On the non-empty subset, CEG-Agent shows stronger composite performance and improves most structure-oriented dimensions. Under Claude, CEG-Claude improves over Claude Code (+Process) on SRE \texttt{error}, \texttt{causal\_edge}, \texttt{main\_chain}, and \texttt{anomaly}, as well as SEE \texttt{error}, \texttt{attached\_to}, and \texttt{causes}. Some exceptions remain, especially textual SRE dimensions such as \texttt{attribution} and the SEE \texttt{contributes\_to} dimension, consistent with the style and edge-coverage effects discussed in \S\ref{app:exp-dim}. Under GPT-5.5, the pattern is stronger: CEG-Codex improves the composite scores and leads most structural and anomaly-related dimensions over Codex (+Process). Overall, removing gold-empty traces clarifies that CEG-Agent's main advantage lies in substantive causal graph construction rather than abstention behavior.

\begin{table*}[t]
\centering
\small
\setlength{\tabcolsep}{2.6pt}
\renewcommand{\arraystretch}{1.12}
\begin{tabularx}{\textwidth}{@{}
>{\raggedright\arraybackslash}p{2.0cm}
>{\raggedright\arraybackslash}p{3.2cm}
c
YYYYYYYY
@{}}
\toprule
\rowcolor{gray!12}
\textbf{Slice} & \textbf{System}
& \cellcolor{blue!14}\textbf{SRE}
& \texttt{error} & \texttt{edge} & \texttt{chain} & \texttt{fail.} & \texttt{attr.} & \texttt{anom.} & \texttt{event} & \texttt{cons.} \\
\midrule
\rowcolor{gray!8}
\multicolumn{11}{@{}l}{\textit{Gold-empty ($n=15$)}} \\
empty & Claude Code (raw)         & \totalcell{64.8} &  6.7 & 100.0 & 100.0 &  6.7 & 100.0 & 37.2 &  80.7 & 100.0 \\
empty & Claude Code (+Process)    & \totalcell{\textbf{87.1}} & \textbf{73.3} & 100.0 & 100.0 & \textbf{73.3} & 100.0 & 51.3 & 100.0 & 100.0 \\
empty & \textbf{CEG-Claude}       & \totalcell{68.6} &  6.7 & 100.0 & 100.0 &  0.0 & 100.0 & \textbf{72.8} & 100.0 &  99.9 \\
\addlinespace[2pt]
empty & Codex (raw)               & \totalcell{59.9} &  0.0 & 100.0 & 100.0 &  0.0 & 100.0 &  0.0 &  98.8 & 100.0 \\
empty & Codex (+Process)          & \totalcell{60.0} &  0.0 & 100.0 & 100.0 &  0.0 & 100.0 &  0.0 & 100.0 & 100.0 \\
empty & \textbf{CEG-Codex}        & \totalcell{\textbf{77.1}} & \textbf{53.3} & 100.0 & 100.0 &  0.0 & 100.0 & \textbf{64.1} & 100.0 & 100.0 \\
\midrule
\rowcolor{gray!8}
\multicolumn{11}{@{}l}{\textit{Non-empty ($n=85$)}} \\
non-empty & Claude Code (raw)      & \totalcell{42.5} & 45.3 & 26.0 & 28.0 & \textbf{51.4} & \textbf{39.6} & 46.0 &  77.7 & 100.0 \\
non-empty & Claude Code (+Process) & \totalcell{43.0} & 45.0 & 31.5 & 27.2 & 43.2 & 34.0 & 45.1 & 100.0 & 100.0 \\
non-empty & \textbf{CEG-Claude}    & \besttotal{48.6} & \textbf{53.8} & \textbf{32.2} & \textbf{42.4} & 45.7 & 26.0 & \textbf{57.7} & 100.0 &  99.9 \\
\addlinespace[2pt]
non-empty & Codex (raw)            & \totalcell{27.1} & 26.9 & 19.8 & 14.6 & {35.2} & {12.4} &  2.4 &  97.8 & 100.0 \\
non-empty & Codex (+Process)       & \totalcell{28.1} & 31.8 & 22.9 & 17.0 & 25.3 & 10.1 &  2.0 & 100.0 & 100.0 \\
non-empty & \textbf{CEG-Codex}     & \besttotal{41.1} & \textbf{44.6} & \textbf{27.1} & \textbf{31.2} & \textbf{41.1} & \textbf{14.8} & \textbf{49.0} & 100.0 & 100.0 \\
\bottomrule
\end{tabularx}
\caption{\textbf{SRE} per-dimension scores split by gold-empty ($n=15$) and non-empty ($n=85$) subsets, with columns aligned to main-paper Table~\ref{tab:exp-sre}. On the empty subset the structural dimensions \texttt{edge}, \texttt{chain}, and \texttt{attr.} are mechanically $100$ because gold and prediction are both empty, and the utility dimensions \texttt{event}/\texttt{cons.} are at or near ceiling for all systems; only \texttt{error}, \texttt{fail.}, and \texttt{anom.} are informative.}
\label{tab:app-exp-empty-sre-dim}
\end{table*}

\begin{table*}[t]
\centering
\small
\setlength{\tabcolsep}{3.5pt}
\renewcommand{\arraystretch}{1.12}
\begin{tabularx}{\textwidth}{@{}
>{\raggedright\arraybackslash}p{2.0cm}
>{\raggedright\arraybackslash}p{3.2cm}
c
YYYYY
@{}}
\toprule
\rowcolor{gray!12}
\textbf{Slice} & \textbf{System}
& \cellcolor{blue!14}\textbf{SEE}
& \texttt{failure} & \texttt{error} & \texttt{attach} & \texttt{cause} & \texttt{contrib.} \\
\midrule
\rowcolor{gray!8}
\multicolumn{8}{@{}l}{\textit{Gold-empty ($n=15$)}} \\
empty & Claude Code (raw)         & \totalcell{0.0} & 0.0 & 0.0 & 0.0 & 0.0 & 0.0 \\
empty & Claude Code (+Process)    & \totalcell{0.0} & 0.0 & 0.0 & 0.0 & 0.0 & 0.0 \\
empty & \textbf{CEG-Claude}       & \totalcell{0.0} & 0.0 & 0.0 & 0.0 & 0.0 & 0.0 \\
\addlinespace[2pt]
empty & Codex (raw)               & \totalcell{0.0} & 0.0 & 0.0 & 0.0 & 0.0 & 0.0 \\
empty & Codex (+Process)          & \totalcell{0.0} & 0.0 & 0.0 & 0.0 & 0.0 & 0.0 \\
empty & \textbf{CEG-Codex}        & \totalcell{0.0} & 0.0 & 0.0 & 0.0 & 0.0 & 0.0 \\
\midrule
\rowcolor{gray!8}
\multicolumn{8}{@{}l}{\textit{Non-empty ($n=85$)}} \\
non-empty & Claude Code (raw)      & \totalcell{16.4} & \textbf{66.0} & 6.2 & 4.6 & 1.2 & 3.8 \\
non-empty & Claude Code (+Process) & \totalcell{22.7} & 55.7 & 21.8 & 19.1 & 1.3 & \textbf{15.5} \\
non-empty & \textbf{CEG-Claude}    & \besttotal{24.1} & 61.3 & \textbf{24.1} & \textbf{20.9} & \textbf{2.4} & 11.7 \\
\addlinespace[2pt]
non-empty & Codex (raw)            & \totalcell{13.7} & {51.6} & 6.6 & 7.0 & 0.0 & 3.6 \\
non-empty & Codex (+Process)       & \totalcell{12.3} & 39.2 & 10.2 & 7.0 & 0.0 & 4.8 \\
non-empty & \textbf{CEG-Codex}     & \besttotal{19.9} & \textbf{56.9} & \textbf{17.9} & \textbf{15.3} & \textbf{1.8} & \textbf{7.5} \\
\bottomrule
\end{tabularx}
\caption{\textbf{SEE} per-dimension scores split by gold-empty ($n=15$) and non-empty ($n=85$) subsets. On the empty subset SEE is $0$ for every system because the gold graph contains no errors, failures, or edges to match.}
\label{tab:app-exp-empty-see-dim}
\end{table*}

\subsection{Graph-Size and Structural Calibration}
\label{app:exp-calibration}

\begin{table*}[t]
\centering
\small
\setlength{\tabcolsep}{3.5pt}
\renewcommand{\arraystretch}{1.12}
\begin{tabularx}{\textwidth}{@{}
>{\raggedright\arraybackslash}p{2.35cm}
>{\raggedright\arraybackslash}p{3.25cm}
YYYYYY
@{}}
\toprule
\rowcolor{gray!12}
\textbf{Backbone} & \textbf{System}
& $\bar{|V_E|}$
& $\bar{|V_R|}$
& $\bar{|V_F|}$
& $\bar{|V_A|}$
& $\bar{|E|}$
& $\bar{|E_C|}$ \\
\midrule
Reference
& Gold
& 34.9 & 3.91 & 1.25 & 2.53 & 45.4 & 7.49 \\
\midrule
Claude-Opus-4.7
& Claude Code (raw)
& 10.7 & 3.14 & 1.02 & 1.20 & 18.2 & 5.30 \\
Claude-Opus-4.7
& Claude Code (+Process)
& 34.9 & 2.09 & 0.74 & 1.24 & 39.5 & 3.56 \\
\rowcolor{blue!3}
Claude-Opus-4.7
& \textbf{CEG-Claude} (ours)
& 34.9 & \textbf{4.69} & \textbf{1.06} & \textbf{2.96} & \textbf{45.4} & \textbf{6.87} \\
\midrule
GPT-5.5
& Codex (raw)
& 34.9 & 1.64 & 1.00 & 0.07 & 37.2 & 1.64 \\
GPT-5.5
& Codex (+Process)
& 34.9 & 1.82 & 1.00 & 0.10 & 37.5 & 1.82 \\
\rowcolor{blue!3}
GPT-5.5
& \textbf{CEG-Codex} (ours)
& 34.9 & \textbf{4.49} & 1.00 & \textbf{2.16} & \textbf{43.0} & \textbf{4.65} \\
\bottomrule
\end{tabularx}
\caption{Mean per-trace graph sizes. Bold marks the closest within-backbone system to the gold reference.}
\label{tab:app-exp-graphsize}
\end{table*}

Table~\ref{tab:app-exp-graphsize} compares each system's output graph size with the gold reference before any matching. The baselines exhibit three calibration failures. First, Claude Code (raw) parses far fewer events than the gold trace representation, which limits downstream graph construction. Second, all baselines under-generate causal structure: Codex variants produce only $1.64$--$1.82$ causal-subgraph edges per trace compared with the gold average of $7.49$, and almost never emit anomalies. Third, even Claude Code (+Process), despite receiving the full event scaffold, under-generates errors, failures, anomalies, and causal edges.

In contrast, both CEG-Agent variants produce graphs much closer to gold scale. CEG-Claude nearly matches the gold total edge count and approaches the gold causal-edge count, while CEG-Codex substantially increases errors, anomalies, and causal edges over the Codex baselines. This calibration explains many of the structural-dimension gains in \S\ref{app:exp-dim}: baselines often fail before matching because they do not propose enough causally relevant nodes and edges, whereas CEG-Agent's incremental construction and validation produce graphs at a scale suitable for causal error reconstruction.

\subsection{Per-Mechanism and Per-Role Recall}
\label{app:exp-mech-role}

\begin{table*}[t]
\centering
\small
\setlength{\tabcolsep}{3.0pt}
\renewcommand{\arraystretch}{1.12}
\begin{tabularx}{\textwidth}{@{}
>{\raggedright\arraybackslash}p{3.15cm}
c
YYYYYY}
\toprule
\rowcolor{gray!12}
\textbf{Slice} & \textbf{$n_{\text{gold}}$}
& \multicolumn{3}{c}{\textbf{Claude-Opus-4.7}}
& \multicolumn{3}{c}{\textbf{GPT-5.5}} \\
\rowcolor{gray!12}
&
& \textbf{CC-raw}
& \textbf{CC-proc}
& \textbf{CEG-Cl.}
& \textbf{Cx-raw}
& \textbf{Cx-proc}
& \textbf{CEG-Cx} \\
\cmidrule(lr){3-5}
\cmidrule(l){6-8}
\rowcolor{gray!8}
\multicolumn{8}{@{}l}{\textit{Mechanism recall on gold $(\varepsilon,\mu)$ keys}} \\
planning             & 77 & 11.7 & 14.3 & \textbf{23.4} & 0.0  & 1.3  & \textbf{7.8}  \\
evidence integration & 72 & 1.4  & 11.1 & \textbf{20.8} & 0.0  & 0.0  & \textbf{13.9} \\
execution            & 66 & 3.0  & 18.2 & \textbf{22.7} & 3.0  & 3.0  & \textbf{12.1} \\
self-evaluation      & 61 & 1.6  & \textbf{23.0} & \textbf{23.0} & 14.8 & 18.0 & \textbf{19.7} \\
control              & 53 & 1.9  & 34.0 & \textbf{35.8} & 24.5 & 20.8 & \textbf{39.6} \\
omission             & 33 & 9.1  & \textbf{18.2} & 15.2 & 0.0  & 0.0  & \textbf{6.1}  \\
representation       & 25 & 0.0  & 12.0 & \textbf{20.0} & 0.0  & 0.0  & 0.0  \\
environment          & 4  & 0.0  & 0.0  & 0.0  & 0.0  & 0.0  & 0.0  \\
\midrule
\rowcolor{gray!8}
\multicolumn{8}{@{}l}{\textit{Structural-role recall on matched errors}} \\
\textsc{root}          & 130 & 10.0 & 21.5 & \textbf{23.1} & 2.3  & 2.3  & \textbf{13.1} \\
\textsc{propagated}    & 155 & 1.9  & 18.7 & \textbf{27.1} & 6.5  & 9.0  & \textbf{16.1} \\
\textsc{amplification} & 106 & 0.9  & 14.2 & \textbf{17.9} & 10.4 & 7.5  & \textbf{16.0} \\
\bottomrule
\end{tabularx}
\caption{Recall (\%) by error mechanism and structural role. Best score within each backbone family is in \textbf{bold}.}
\label{tab:app-exp-mech-role}
\end{table*}

Table~\ref{tab:app-exp-mech-role} shows where CEG-Agent improves error recovery within the controlled vocabulary. Both CEG-Agent variants achieve the best within-backbone recall on most mechanisms and on all three structural roles. The largest gains occur on cumulative failure modes such as \emph{representation}, \emph{evidence integration}, and \emph{planning}, which require aggregating evidence across multiple trace steps rather than detecting a single local mistake.

The Claude-family results show that CEG-Claude improves recall on nearly every mechanism, with the main exception of \emph{omission}, where Claude Code (+Process) is slightly stronger, and \emph{environment}, which has only four examples and is missed by all systems. Under GPT-5.5, the contrast is sharper: Codex baselines are near zero on several mechanisms, while CEG-Codex recovers non-trivial recall on \emph{planning}, \emph{evidence integration}, \emph{execution}, \emph{control}, and \emph{omission}. The role-level results are particularly revealing: CEG-Agent's largest gains are on \textsc{propagated} errors, which require identifying downstream errors together with their upstream causal predecessors. This supports the central role of the typed causal subgraph $G_C$ in recovering error propagation rather than only isolated root causes.

\subsection{Failure-Type Coverage}
\label{app:exp-failuretype}

\begin{table*}[t]
\centering
\small
\setlength{\tabcolsep}{3.5pt}
\renewcommand{\arraystretch}{1.12}
\begin{tabularx}{\textwidth}{@{}
>{\raggedright\arraybackslash}p{3.25cm}
YYYYYYY
@{}}
\toprule
\rowcolor{gray!12}
\textbf{System}
& \textbf{Completion}
& \textbf{Correct.}
& \textbf{Constraint}
& \textbf{Efficiency}
& \textbf{Safety}
& \textbf{Other}
& \textbf{Coverage} \\
\midrule
Gold
& 53 & 43 & 17 & 6 & 3 & 3 & 6/6 \\
\midrule
\rowcolor{gray!8}
\multicolumn{8}{@{}l}{\textit{Claude-Opus-4.7}} \\
Claude Code (raw)
& 62 & 21 & 19 & 0 & 0 & 0 & 3/6 \\
Claude Code (+Process)
& 49 & 14 & 10 & 1 & 0 & 0 & 4/6 \\
\rowcolor{blue!3}
\textbf{CEG-Claude} (ours)
& 42 & 33 & 21 & 5 & 0 & 5 & \textbf{5/6} \\
\midrule
\rowcolor{gray!8}
\multicolumn{8}{@{}l}{\textit{GPT-5.5}} \\
Codex (raw)
& 40 & 33 & 25 & 0 & 0 & 2 & 4/6 \\
Codex (+Process)
& 35 & 59 & 0 & 6 & 0 & 0 & 3/6 \\
\rowcolor{blue!3}
\textbf{CEG-Codex} (ours)
& 28 & 50 & 6 & 0 & 0 & 16 & 4/6 \\
\bottomrule
\end{tabularx}
\caption{Failure-type emission counts across 100 traces. Coverage is the number of non-empty failure types emitted.}
\label{tab:app-exp-failure-type}
\end{table*}

Table~\ref{tab:app-exp-failure-type} summarizes failure-type coverage. CEG-Claude covers five of the six failure categories, including the rare \emph{efficiency} and \emph{other} buckets that most baselines ignore. CEG-Codex also broadens the GPT-5.5 emission distribution, partially correcting the degenerate behavior of Codex (+Process), which emits no \emph{constraint} failures across the benchmark.
Three patterns stand out. First, most systems overuse \emph{completion} as a default failure type when uncertain, while CEG-Agent is less concentrated in this bucket. Second, \emph{safety} failures are missed by all systems, despite three gold instances, suggesting that safety-specific diagnosis may require an explicit policy or oracle beyond trace-only evidence. Third, CEG-Agent generally produces a more diverse failure-type distribution, which is consistent with its explicit separation of outcome-level failures from process-level errors.

\subsection{Repair-Value Calibration}
\label{app:exp-repair}

Repair-value prediction is one of the most difficult components of the pipeline. Directly comparing predicted buckets with gold buckets is affected by severe class imbalance: the gold distribution is long-tailed, with only $3.3\%$ of errors labeled \textsc{high}, so predicting \textsc{low} for most errors can obtain deceptively high bucket accuracy. For downstream debugging, a more useful question is whether a system can \emph{rank} errors by repair priority within a trace, since repair values are primarily intended to help decide which error to fix first. We therefore report both absolute bucket confusion and relative ranking quality in Tables~\ref{tab:app-exp-repair-conf} and \ref{tab:app-exp-repair-rank}.

\begin{table*}[t]
\centering
\small
\setlength{\tabcolsep}{4.0pt}
\renewcommand{\arraystretch}{1.12}
\begin{tabularx}{\textwidth}{@{}
>{\raggedright\arraybackslash}p{2.2cm}
YYY
YYY}
\toprule
\rowcolor{gray!12}
\textbf{Gold}
& \multicolumn{3}{c}{\textbf{CEG-Claude prediction}}
& \multicolumn{3}{c}{\textbf{CEG-Codex prediction}} \\
\rowcolor{gray!12}
& \textsc{low} & \textsc{medium} & \textsc{high}
& \textsc{low} & \textsc{medium} & \textsc{high} \\
\cmidrule(lr){2-4}
\cmidrule(l){5-7}
\textsc{low}
& \cellcolor{blue!6}\textbf{34} & 13 & 5
& \cellcolor{blue!6}\textbf{39} & 0 & 0 \\
\textsc{medium}
& 14 & \cellcolor{blue!6}\textbf{7} & 9
& 17 & \cellcolor{blue!6}\textbf{0} & 0 \\
\textsc{high}
& 1 & 1 & \cellcolor{blue!6}\textbf{0}
& 0 & 0 & \cellcolor{blue!6}\textbf{1} \\
\midrule
\textbf{Column total}
& 49 & 21 & 14
& 56 & 0 & 1 \\
\bottomrule
\end{tabularx}
\caption{Absolute repair-value confusion on $(\varepsilon,\mu)$-matched errors. Diagonal cells indicate exact bucket agreement.}
\label{tab:app-exp-repair-conf}
\end{table*}

\begin{table*}[t]
\centering
\small
\setlength{\tabcolsep}{5.0pt}
\renewcommand{\arraystretch}{1.12}
\begin{tabularx}{\textwidth}{@{}
>{\raggedright\arraybackslash}p{3.6cm}
YYYY
@{}}
\toprule
\rowcolor{gray!12}
\textbf{System} & \textbf{Matched errors}
& \textbf{Pairwise agree.} & \textbf{Kendall's $\tau$}
& \textbf{Bucket acc.} \\
\midrule
Claude Code (+Process)   & 65 & 0.528 & $+0.056$ & 0.062 \\
Codex (+Process)         & 23 & 0.500 & $\phantom{+}0.000$ & 0.000 \\
\rowcolor{blue!3}
\textbf{CEG-Claude}      & 84 & \textbf{0.700} & $\boldsymbol{+0.400}$ & 0.488 \\
\rowcolor{blue!3}
\textbf{CEG-Codex}       & 57 & 0.500 & $\phantom{+}0.000$ & \textbf{0.702} \\
\bottomrule
\end{tabularx}
\caption{Relative repair-value ranking on $(\varepsilon,\mu)$-matched errors. Pairwise agreement measures whether predicted priorities preserve the gold ordering within each trace.}
\label{tab:app-exp-repair-rank}
\end{table*}

\paragraph{Relative ranking is more informative than absolute buckets.}
Table~\ref{tab:app-exp-repair-rank} evaluates whether predicted repair values preserve the relative priority ordering of gold errors within the same trace. This view is more informative than absolute bucket accuracy. Codex (+Process) and CEG-Codex both have Kendall's $\tau=0.0$, indicating that their predictions carry little priority-ordering information, even though CEG-Codex achieves the highest bucket accuracy by predicting almost all errors as \textsc{low}. In contrast, CEG-Claude achieves substantially better pairwise agreement and a positive Kendall's $\tau=+0.400$, suggesting that it captures some useful within-trace priority structure even when its absolute bucket calibration is imperfect.

\paragraph{Backbone-specific calibration patterns.}
The two CEG-Agent variants fail in different ways. CEG-Claude over-predicts \textsc{high}: $20.0\%$ of its errors are assigned \textsc{high}, compared with a gold prevalence of $3.3\%$. However, its positive ranking score suggests that its underlying severity ordering is partially useful, and that the main problem is bucket-threshold calibration. CEG-Codex shows the opposite failure mode: it collapses almost entirely to \textsc{low}, never emits \textsc{medium}, and therefore obtains high absolute accuracy without providing meaningful repair prioritization. Thus, bucket accuracy alone is misleading for repair-value evaluation.

\paragraph{Source of calibration error.}
The calibration error mainly comes from two signals that are sensitive to LLM writing style: the counterfactual severance signal $\sigma(r,f^\star)$, which depends on how the model describes causal bottlenecks, and the lexical hedging signal $h(r)$, which varies with confident or cautious wording. Future work may compute $\sigma(r,f^\star)$ deterministically from the graph, constrain the influence of $h(r)$, or tune bucket thresholds for specific downstream applications. We do not further study fine-grained calibration here, since CEG-Claude already provides useful relative ordering in the current evaluation.

\subsection{Inter-System Agreement and Irreducibly Hard Traces}
\label{app:exp-agreement}

\begin{table*}[t]
\centering
\small
\setlength{\tabcolsep}{5.0pt}
\renewcommand{\arraystretch}{1.12}
\begin{tabularx}{\textwidth}{@{}
>{\raggedright\arraybackslash}p{5.0cm}
YYYY
@{}}
\toprule
\rowcolor{gray!12}
\textbf{Cohort} & \textbf{$n$} & \textbf{Tau2Bench} & \textbf{SWE-bench} & \textbf{BrowseComp} \\
\midrule
All systems have SEE $=0$ & 19 & 10 & 3 & 6 \\
\quad Gold-empty subset & 15 & 10 & 3 & 2 \\
\quad Non-empty all-fail subset & 4 & 0 & 0 & 4 \\
\bottomrule
\end{tabularx}
\caption{Traces on which all six systems obtain SEE $=0$, decomposed by source.}
\label{tab:app-exp-agreement}
\end{table*}

We further examine inter-system agreement by pooling per-trace SEE scores from all six evaluated systems. As shown in Table~\ref{tab:app-exp-agreement}, there are 19 traces on which every system obtains SEE $=0$. Most of these cases are the gold-empty traces analyzed in \S\ref{app:exp-empty}, where SEE becomes degenerate because the gold graph contains no causal structure to recover. The remaining four cases are non-empty BrowseComp traces, forming a small irreducibly hard cohort for the current systems.
Manual inspection of these four cases suggests a common pattern: the failure often arises from global evidence integration across a long research trajectory, rather than from a single locally identifiable error event. Such cases are especially difficult for structurally exact evaluation because the relevant causal evidence is distributed across multiple evidence-gathering and reasoning steps. This explains why even strong systems fail to recover the exact annotated structure on these traces.
At the other end of the spectrum, the easiest traces are mostly from Tau2Bench: among the 10 traces with maximum SEE at least $0.58$ under some system, 8 come from Tau2Bench. This source-level contrast reinforces the broader pattern in \S\ref{app:exp-source}: Tau2Bench is structurally easier and more affected by abstention behavior, whereas SWE-bench and BrowseComp require richer causal reconstruction and better expose the benefits of CEG-Agent.

\subsection{Per-Backbone Diagnostic Capability Profile}
\label{app:exp-backbone-detail}

This subsection expands \S\ref{sec:exp-backbone} with a per-dimension breakdown of CEG-Agent under five LLM backbones: Claude-Opus-4.7, GPT-5.5, Gemini-3.1-Pro, Qwen3.7-Max, and GLM-5.1. The schema, tool set, and inference budget are fixed across backbones, so the differences below reflect backbone-specific diagnostic behavior under the same CEG-Agent framework. Tables~\ref{tab:app-exp-backbone-sre} and \ref{tab:app-exp-backbone-see} report the SRE and SEE breakdowns.

\begin{table*}[t]
\centering
\small
\setlength{\tabcolsep}{3.5pt}
\renewcommand{\arraystretch}{1.12}
\begin{tabularx}{\textwidth}{@{}
>{\raggedright\arraybackslash}p{3.0cm}
c
YYYYYYY
@{}}
\toprule
\rowcolor{gray!12}
\textbf{CEG-Agent backbone}
& \cellcolor{blue!14}\textbf{SRE}
& \multicolumn{3}{c}{\textbf{Structural}}
& \multicolumn{3}{c}{\textbf{Textual}} \\
\rowcolor{gray!12}
&
& \texttt{error} & \texttt{edge} & \texttt{chain}
& \texttt{fail.} & \texttt{attr.} & \texttt{anom.} \\
\cmidrule(lr){2-2}
\cmidrule(lr){3-5}
\cmidrule(l){6-8}
Claude-Opus-4.7  & \besttotal{51.6} & \textbf{46.7} & \textbf{42.4} & \textbf{51.0} & 38.8          & \textbf{37.1} & \textbf{60.0} \\
GLM-5.1          & \totalcell{46.7} & 45.1          & 38.0          & 39.7          & \textbf{41.9} & 37.0          & 42.8 \\
GPT-5.5          & \totalcell{46.5} & 45.9          & 38.0          & 41.5          & 34.9          & 27.6          & 51.3 \\
Gemini-3.1-Pro   & \totalcell{44.2} & 40.8          & 41.0          & 37.3          & 35.7          & 26.5          & 43.1 \\
Qwen3.7-Max      & \totalcell{43.7} & 41.7          & 37.8          & 37.8          & 34.2          & 32.8          & 35.2 \\
\bottomrule
\end{tabularx}
\caption{Per-dimension \textbf{SRE} scores of CEG-Agent under five LLM backbones. Best score per column is in \textbf{bold}. Utility dimensions (\texttt{event}, \texttt{cons.}) are at ceiling for all backbones and are omitted; the composite SRE in the second column already incorporates them.}
\label{tab:app-exp-backbone-sre}
\end{table*}

\begin{table*}[t]
\centering
\small
\setlength{\tabcolsep}{4.0pt}
\renewcommand{\arraystretch}{1.12}
\begin{tabularx}{\textwidth}{@{}
>{\raggedright\arraybackslash}p{3.0cm}
c
YYYYY
@{}}
\toprule
\rowcolor{gray!12}
\textbf{CEG-Agent backbone}
& \cellcolor{blue!14}\textbf{SEE}
& \multicolumn{5}{c}{\textbf{Dimensions}} \\
\rowcolor{gray!12}
&
& \texttt{failure} & \texttt{error} & \texttt{attach} & \texttt{cause} & \texttt{contrib.} \\
\cmidrule(lr){2-2}
\cmidrule(l){3-7}
Claude-Opus-4.7  & \besttotal{20.5} & 52.1          & \textbf{20.5} & \textbf{17.8} & 2.0          & 9.9          \\
GLM-5.1          & \totalcell{19.6} & \textbf{54.9} & 18.0          & 15.0          & 1.3          & 8.7          \\
Qwen3.7-Max      & \totalcell{18.7} & 46.1          & 18.9          & 16.1          & \textbf{2.3} & \textbf{10.2} \\
GPT-5.5          & \totalcell{16.9} & 48.3          & 15.2          & 13.0          & 1.5          & 6.3          \\
Gemini-3.1-Pro   & \totalcell{15.0} & 42.0          & 15.5          & 11.4          & 1.3          & 4.8          \\
\bottomrule
\end{tabularx}
\caption{Per-dimension \textbf{SEE} scores of CEG-Agent under five LLM backbones. Best score per column is in \textbf{bold}.}
\label{tab:app-exp-backbone-see}
\end{table*}

\begin{table*}[t]
\centering
\small
\setlength{\tabcolsep}{4.0pt}
\renewcommand{\arraystretch}{1.12}
\begin{tabularx}{\textwidth}{@{}
>{\raggedright\arraybackslash}p{2.0cm}
YYYYY
@{}}
\toprule
\rowcolor{gray!12}
\textbf{Metric}
& \textbf{CEG-Claude}
& \textbf{CEG-Codex}
& \textbf{CEG-Gemini}
& \textbf{CEG-Qwen}
& \textbf{CEG-GLM} \\
\midrule
\rowcolor{gray!8}
\multicolumn{6}{@{}l}{\textit{Mean per trace}} \\
LLM calls            & 18.7      & 19.0       & 23.5       & 23.2       & 24.1       \\
Agent iterations     & 13.5      & 14.0       & 18.2       & 18.0       & 18.8       \\
Input tokens         & 556{,}039 & 350{,}076  & 467{,}519  & 622{,}982  & 436{,}678  \\
Output tokens        & 3{,}832   & 4{,}832    & 12{,}567   & 19{,}063   & 20{,}332   \\
Wall-clock time (s)  & 95.0      & 160.0      & 372.6      & 214.4      & 517.5      \\
\bottomrule
\end{tabularx}
\caption{Cost profile of CEG-Agent under five LLM backbones on CEG-Bench.}
\label{tab:app-exp-cost}
\end{table*}

\begin{table*}[t]
\centering
\small
\setlength{\tabcolsep}{4.0pt}
\renewcommand{\arraystretch}{1.12}
\begin{tabularx}{\textwidth}{@{}
>{\raggedright\arraybackslash}p{4.0cm}
YYYYY
@{}}
\toprule
\rowcolor{gray!12}
\textbf{Tool}
& \textbf{CEG-Claude}
& \textbf{CEG-Codex}
& \textbf{CEG-Gemini}
& \textbf{CEG-Qwen}
& \textbf{CEG-GLM} \\
\midrule
\texttt{find\_failures}           & 108 & 101 &  99 & 106 & 101 \\
\texttt{propose\_errors}          & 102 & 100 &  96 & 124 & 110 \\
\texttt{find\_anomalies}          & 101 & 100 &  94 & 103 & 100 \\
\texttt{build\_causal\_graph}     & 102 & 100 &  87 & 106 & 100 \\
\texttt{estimate\_repair\_values} & 100 & 101 &  88 &  96 &  90 \\
\texttt{validate\_graph}          & 112 & 136 &  98 & 111 & 121 \\
\texttt{apply\_critic\_patch}     &  41 &  71 & 103 &  74 &  94 \\
\texttt{spawn\_subagent}          &   5 &  13 &  32 &   6 &   9 \\
\bottomrule
\end{tabularx}
\caption{Total agent-issued tool invocations across the evaluated traces for each backbone. Infrastructure tools (\texttt{read\_trace}, \texttt{build\_events}, \texttt{render\_graph}, \texttt{finalize}) and scaffolding tools (\texttt{write\_todo}, \texttt{inspect\_state}) are omitted from the table, since they are not central to the tool-use analysis.}
\label{tab:app-exp-tools}
\end{table*}

\paragraph{Claude-Opus-4.7 is the most balanced diagnostician.}
Claude achieves the highest composite SRE and SEE scores, leads most SRE structural dimensions, and has no clear weak dimension relative to the other backbones. Its largest advantage is on SRE \texttt{anomaly} ($60.0$), indicating reliable use of the anomaly-detection component. It also achieves the strongest SRE \texttt{main\_chain} score ($51.0$), suggesting that it is particularly effective at organizing local errors into a coherent causal propagation chain.

\paragraph{GLM-5.1 is strong on exact reconstruction and failure typing.}
GLM-5.1 is the strongest non-Claude backbone on SEE and is competitive with GPT-5.5 on SRE. It leads all backbones on SRE \texttt{failure} ($41.9$) and SEE \texttt{failure} ($54.9$), and nearly matches Claude on SRE \texttt{attribution}. Its main weaknesses are SRE \texttt{anomaly} and \texttt{main\_chain}, both of which remain below Claude. The cost and tool-use profile in \S\ref{app:exp-cost} suggests that GLM relies more heavily on validator and critic feedback, consistent with a model that can recover strong exact structure after schema violations are surfaced. Overall, GLM appears strong at identifying failure categories and exact structure, but less effective at recovering the broader anomaly and propagation-chain context.

\paragraph{GPT-5.5 is competitive on error identification but weaker on structure.}
GPT-5.5 closely tracks Claude on SRE \texttt{error} ($45.9$ vs.\ $46.7$), showing that it can identify individual process-level errors reasonably well. However, it loses ground on SRE \texttt{main\_chain} and \texttt{attribution}, and also trails Claude on SEE \texttt{error}, \texttt{attached\_to}, and \texttt{contributes\_to}. This indicates that GPT-5.5 is better at describing local errors than at wiring them into an exact causal structure.

\paragraph{Gemini-3.1-Pro is balanced on causal edges but conservative on contribution.}
Gemini achieves a competitive SRE \texttt{edge} score ($41.0$), close to Claude, but obtains the lowest SEE \texttt{contributes\_to} score. This suggests that Gemini often captures local causal links between errors, but is more conservative or incomplete when linking errors to outcome-level failures. Its lower SEE \texttt{failure} score also indicates weaker alignment with the controlled failure-type vocabulary.

\paragraph{Qwen3.7-Max suppresses anomalies but performs well on exact causal links.}
Qwen has the lowest SRE \texttt{anomaly} score ($35.2$), suggesting a tendency to collapse surface irregularities into process-level errors rather than emitting separate anomaly nodes. However, it performs strongly on exact structural dimensions: it achieves the best SEE \texttt{cause} and \texttt{contributes\_to} scores, and ranks third overall on SEE. Thus, although Qwen has lower semantic similarity, its predicted causal links are often well formed under exact tuple matching.

\paragraph{SRE and SEE capture different capabilities.}
The SRE and SEE rankings diverge beyond the top model. In particular, Qwen3.7-Max outperforms both GPT-5.5 and Gemini-3.1-Pro on structurally exact reconstruction despite lower semantic similarity. This shows that fluent diagnostic descriptions and exact causal graph reconstruction are not the same capability. More broadly, composite scores can hide important differences across dimensions: backbones with similar overall scores may fail in different ways. This supports reporting both per-dimension and per-backbone results when evaluating systems on CEG-Bench.

\paragraph{Implications for backbone choice.}
These results suggest that backbone choice should depend on the intended use of the CEG. Claude-Opus-4.7 is the strongest overall option, especially when both semantic quality and structural fidelity matter. GLM-5.1 is a strong non-Claude alternative for exact reconstruction and failure typing. GPT-5.5 remains competitive on local error identification but is weaker on propagation structure. Gemini is better suited when local causal-edge recovery is more important than failure-link completeness, while Qwen is useful when exact causal links matter more than anomaly coverage. Overall, CEG-Agent is not tied to a single LLM family, and CEG-Bench exposes meaningful backbone-specific diagnostic profiles that are obscured by aggregate scores alone.

\subsection{Cost and Tool-Use Profile}
\label{app:exp-cost}

Tables~\ref{tab:app-exp-cost} and \ref{tab:app-exp-tools} summarize the cost and tool-use profile of CEG-Agent across five backbones. Cost varies substantially by backbone. Claude-Opus-4.7 and GPT-5.5 converge in about 13--14 iterations with roughly 19 LLM calls per trace, while Gemini, Qwen, and GLM require more iterations and approach the budget more often on long traces. Across all backbones, input tokens dominate output tokens because the normalized trace is repeatedly attached during iterative diagnosis. Wall-clock time is governed not only by token count but also by provider latency: GLM is the slowest despite moderate input-token usage, while Qwen uses the most input tokens but has lower latency than GLM.

Tool-use patterns are stable for the core diagnosis tools: all backbones call \texttt{find\_failures}, \texttt{propose\_errors}, \texttt{find\_anomalies}, \texttt{build\_causal\_graph}, and \texttt{estimate\_repair\_values} at roughly one call per trace. The main differences appear in optional refinement behavior. \texttt{apply\_critic\_patch} is used much more often by Gemini and GLM than by Claude, indicating that these backbones rely more heavily on critic-guided repair. \texttt{spawn\_subagent} is most frequent under Gemini, suggesting greater delegation on difficult traces. Overall, stronger backbones reach high-quality CEGs with fewer repair and delegation steps, while less stable backbones make greater use of CEG-Agent's validator and critic loop to recover from imperfect first drafts.


\paragraph{End-to-end monetary cost.}
CEG-Agent performs diagnosis as a one-time offline procedure for each failed trace, so its cost should be compared with running the underlying coding agent on the same trace. Under the Claude-Opus-4.7 backbone, the total billed cost over the full benchmark is \$300 for Claude Code (raw) and \$244 for Claude Code (+Process), whereas \textbf{CEG-Claude costs \$225}. Thus, the iterative diagnostic procedure does not increase end-to-end cost relative to the coding-agent baselines. The main reason is that the normalized trace is reused across diagnostic steps and is therefore cache-friendly. In addition, CEG-Agent emits compact structured graph updates, averaging about $3.8$k output tokens per trace, while free-form baselines spend more tokens on unstructured exploration and long diagnostic reports.

\section{Closed-Loop Repair Study}
\label{app:exp-optimization}

This appendix provides additional details for the closed-loop repair probe in \S\ref{sec:exp-optimization}. The goal of this study is not to claim a fully optimized repair system, but to test whether CEGs contain actionable information that can help a model recover from a previously failed execution.

\paragraph{Protocol.}
All 100 traces in CEG-Bench are failed attempts, so the original runs have zero success by construction and are omitted from the tables. For each trace, we re-attempt the original task once with a feedback hint derived from the AAAP annotation. We compare a content-free \textbf{generic} retry note with three compact CEG-based hints: \textbf{CEG: failure}, which exposes only the outcome-level failure; \textbf{CEG: errors}, which exposes the causal error nodes with their mechanisms, roles, and descriptions; and \textbf{CEG: priority errors}, which exposes only errors with high or medium repair value. The repair-value filter is intended to test whether the counterfactual repair signal can identify a smaller and more actionable subset of errors.

The backbone is fixed to GLM-4.6 with temperature $0$. On \textbf{Tau2Bench} ($n=36$), the model is re-run as the agent in the real Tau2 simulator with the feedback injected into the domain policy, and success is measured by the task reward. On \textbf{BrowseComp} ($n=29$), the model re-derives an answer from the failed trace and the feedback without new search, and the answer is judged by the BrowseComp/HLE grader. On \textbf{SWE-bench} ($n=35$), the model generates an oracle-file patch from the issue and gold-patch-touched files at \texttt{base\_commit}, and the patch is graded by the official test-based harness.

\begin{table}[t]
\small
\centering
\setlength{\tabcolsep}{4pt}
\renewcommand{\arraystretch}{1.08}
\begin{tabularx}{\columnwidth}{l>{\centering\arraybackslash}X>{\centering\arraybackslash}X>{\centering\arraybackslash}X}
\toprule
\rowcolor{gray!15}
\textbf{Feedback} & \textbf{Tau2} & \textbf{Browse} & \textbf{SWE} \\
\midrule
generic              & 77.8 & 6.9  & 14.3 \\
\midrule
CEG: failure         & \textbf{91.7} & 10.3 & \textbf{17.1} \\
CEG: errors          & 80.6 & \textbf{17.2} & \textbf{17.1} \\
CEG: priority errors & 83.3 & \textbf{17.2} & 8.6 \\
\bottomrule
\end{tabularx}
\caption{Closed-loop repair pass@1 (\%) under the feedback conditions used in the main paper.}
\label{tab:app-opt-main}
\end{table}

\paragraph{Overall results.}
Table~\ref{tab:app-opt-main} shows that CEG feedback is most useful when the agent can act on it interactively. On Tau2Bench, all CEG-derived hints outperform the generic retry, and the failure-level hint gives the largest gain ($91.7$ vs.\ $77.8$). This indicates that naming the outcome-level failure provides a concrete correction target for re-execution. The two error-based hints also improve over generic feedback, suggesting that process-level causal errors can guide the agent toward more appropriate next actions.

The pattern differs in the two single-shot settings. On BrowseComp, error-based hints achieve the best score ($17.2$), while the failure-only hint is weaker ($10.3$). This is consistent with the nature of deep-research tasks: merely knowing that the final answer is wrong is often insufficient, whereas causal errors can point the model toward reasoning or evidence-integration mistakes. However, the absolute success rate remains low because the setting does not allow new retrieval; if the failed trace never contains the missing evidence, the answer often cannot be recovered by re-reading the trace alone. On SWE-bench, concise failure and error hints slightly improve over generic feedback, but priority-error filtering hurts. This suggests that the current repair-value signal is not yet robust enough for single-shot code patching, where filtering out medium-context errors may remove information needed to produce an applicable and complete patch.

\begin{table}[t]
\small
\centering
\setlength{\tabcolsep}{3pt}
\renewcommand{\arraystretch}{1.08}
\begin{tabularx}{\columnwidth}{l>{\centering\arraybackslash}X>{\centering\arraybackslash}X>{\centering\arraybackslash}X>{\centering\arraybackslash}X>{\centering\arraybackslash}X}
\toprule
\rowcolor{gray!15}
\textbf{Feedback} & \textbf{pass@1} & \textbf{apply-fail} & \textbf{test-fail} & \textbf{churn} & \textbf{test\%} \\
\midrule
generic              & 14.3 & 13 & 17 & 5.6  & 9  \\
CEG: failure         & \textbf{17.1} & 19 & 10 & 16.5 & 17 \\
CEG: errors          & \textbf{17.1} & 14 & 15 & 9.7  & 14 \\
CEG: priority errors & 8.6  & 19 & 13 & 10.0 & 9  \\
\bottomrule
\end{tabularx}
\caption{SWE-bench closed-loop breakdown ($n=35$). \texttt{apply-fail} counts patches that fail to apply, \texttt{test-fail} counts patches that apply but fail tests, \texttt{churn} is the mean number of changed lines, and \texttt{test\%} is the percentage of patches that edit test files.}
\label{tab:app-opt-swe}
\end{table}

\paragraph{SWE-bench patch behavior.}
Table~\ref{tab:app-opt-swe} explains why CEG feedback is less stable for single-shot code repair. The concise failure and error hints each solve one additional instance over the generic retry, but they do so through different behaviors. The error hint keeps patch complexity moderate and has a lower apply-failure count than the failure-only hint. The failure-only hint induces larger patches and more test-file edits, which increases apply failures but also reduces the number of apply-but-fail-test cases among the patches that do apply. In contrast, priority-error filtering performs worse than generic feedback. Although it is intended to focus the model on the most repair-relevant errors, it can remove contextual errors that are necessary for constructing a correct patch, leading the model to over-focus on an incomplete local fix.

These results highlight an important distinction between diagnosis and repair. A CEG can identify causally relevant errors and rank their likely repair value, but the best way to expose this information to a downstream agent depends on the execution setting. Interactive agents can use a compact failure or error signal to adjust future actions, while single-shot patch generators may require broader context to avoid brittle or incomplete edits. Thus, CEGs provide useful repair information, but feedback formatting and repair-value calibration remain open design problems.

\paragraph{Scope.}
The closed-loop study is intentionally small, with $29$--$36$ traces per source benchmark, so the results should be interpreted as directional rather than statistically decisive. Its main value is to show that CEGs are not only evaluative artifacts: they can also function as structured feedback for downstream recovery. The strongest and most consistent gains appear in the interactive Tau2Bench setting, where the agent can directly act on the diagnostic signal. We leave larger-scale closed-loop optimization, better feedback formatting, and more robust repair-value calibration to future work.

\subsection{Case Studies}
\label{app:exp-case}

\begin{figure*}[t]
\centering
\includegraphics[width=\linewidth]{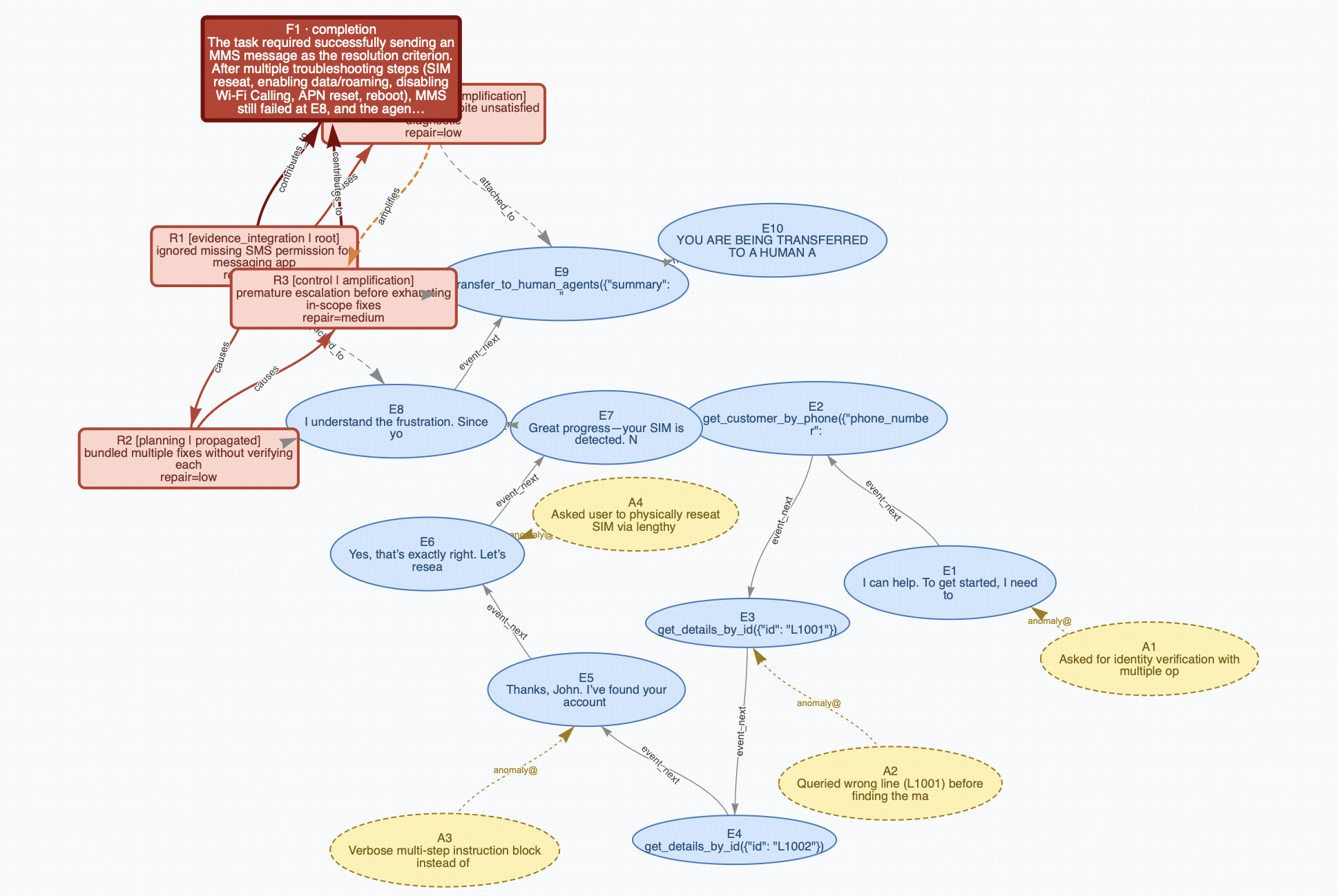}
\caption{CEG-Claude visualization for a Tau2Bench trace.}
\label{fig:case-tau2}
\end{figure*}

\begin{figure*}[t]
\centering
\includegraphics[width=\linewidth]{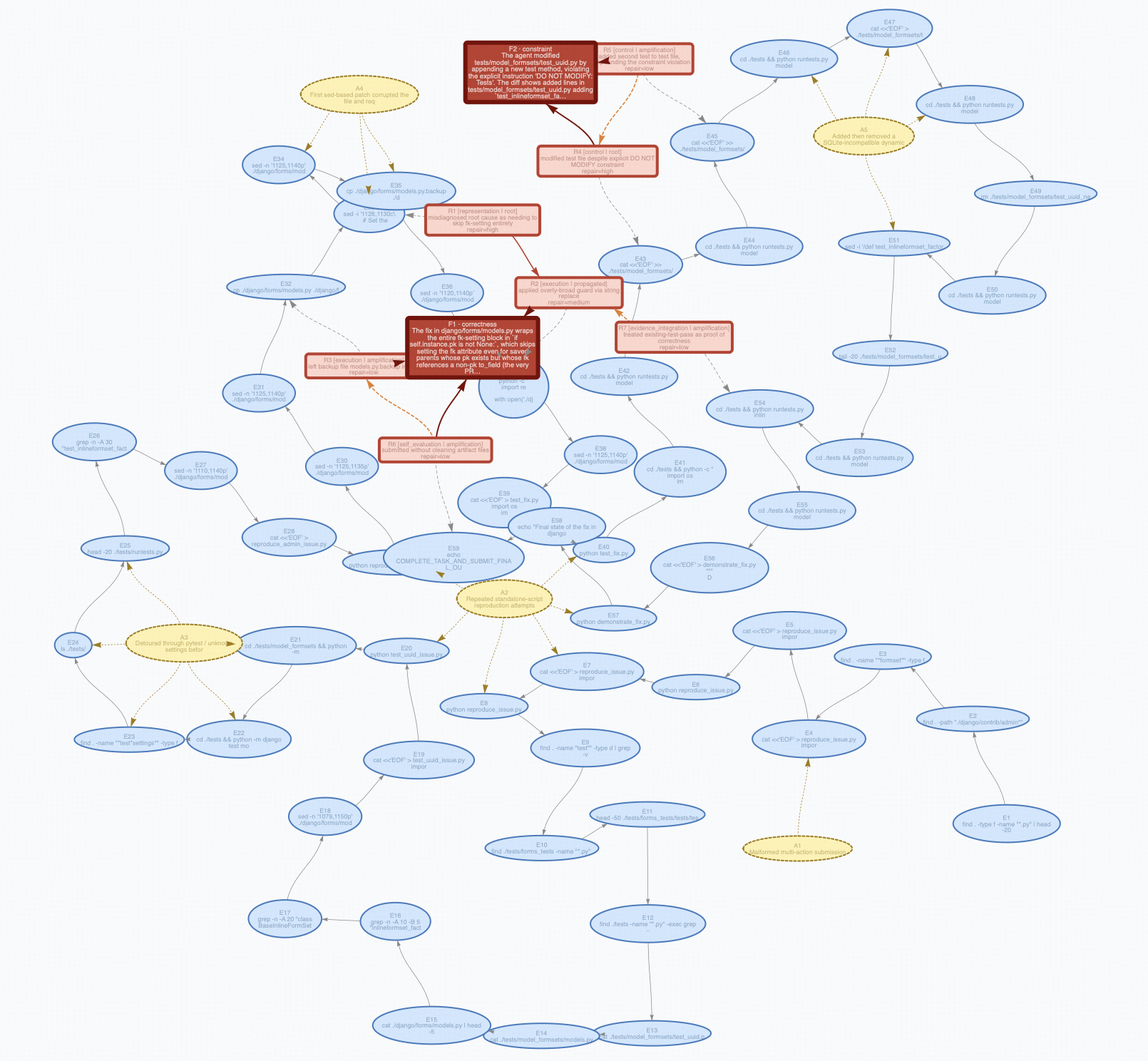}
\caption{CEG-Claude visualization for a SWE-bench trace.}
\label{fig:case-swe}
\end{figure*}

\begin{figure*}[t]
\centering
\includegraphics[width=\linewidth]{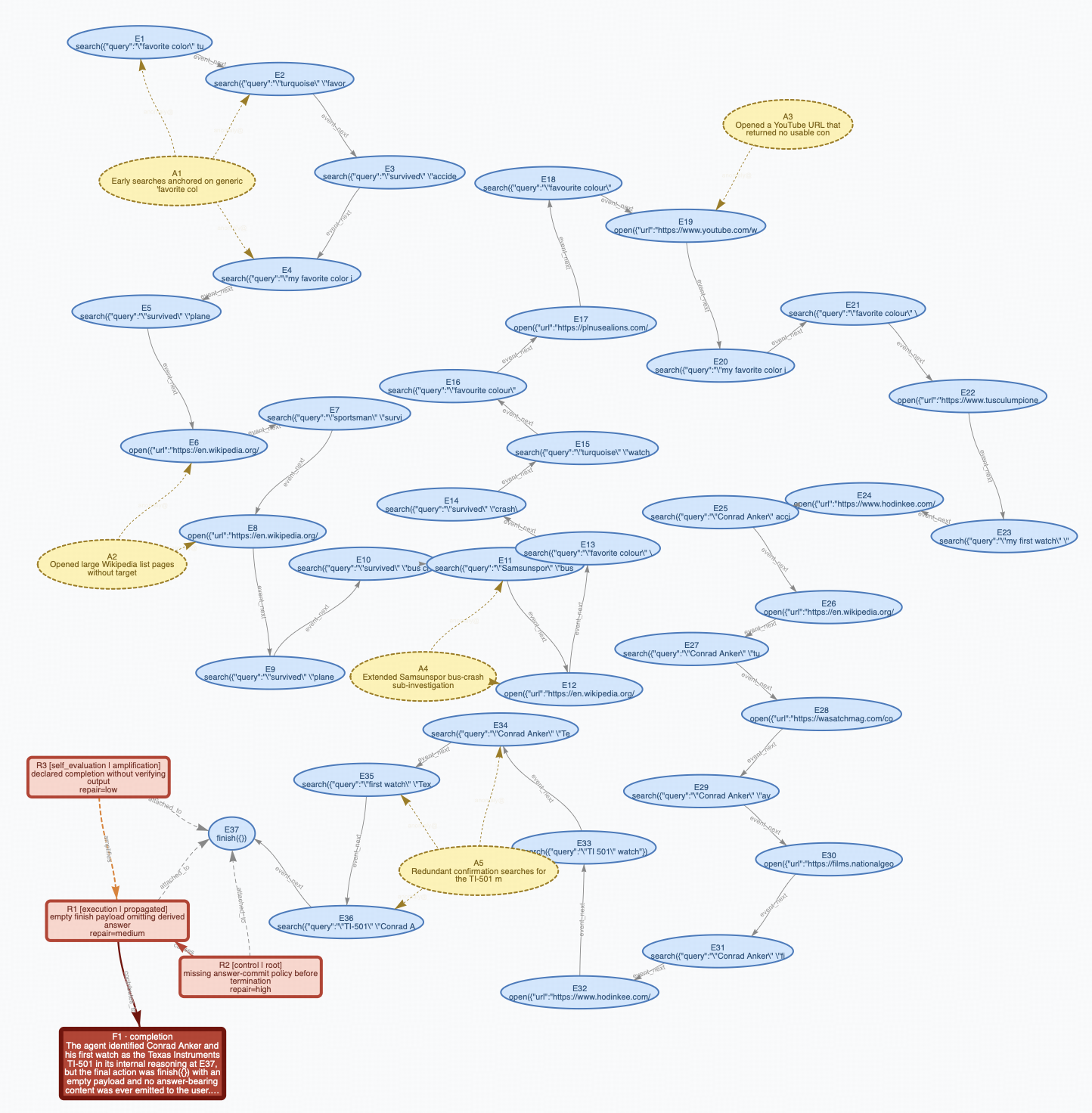}
\caption{CEG-Claude visualization for a BrowseComp trace.}
\label{fig:case-browse}
\end{figure*}

Figures~\ref{fig:case-tau2}--\ref{fig:case-browse} show representative HTML visualizations produced by CEG-Claude on traces from Tau2Bench, SWE-bench, and BrowseComp. Each visualization aligns the chronological event sequence with the induced causal error graph, making errors, anomalies, propagation chains, and outcome-level failures inspectable in a single view. These examples illustrate that CEG-Agent produces not only machine-evaluable CEGs, but also human-readable diagnostic artifacts for debugging long agentic traces.

\section{Limitations}
\label{sec:limitations}

This work has several limitations. First, the released CEG-Bench annotations are produced automatically by AAAP rather than by exhaustive human annotation. We validate them against an 8-expert human gold set (CEG-Bench-Gold, \S\ref{app:human-gold}), which agrees closely with the automatic annotations ($\mathrm{CEG\text{-}Sim}=71.5$; error/failure/anomaly $76.3/76.3/88.1$), but the AAAP annotations may still contain systematic biases shared by the agentic annotators. In particular, some failed traces are retained with empty CEGs because no clear causal error can be inferred from the trace alone without external outcome feedback. We make this choice to preserve the generality and reproducibility of CEG construction, but acknowledge that human experts might adjudicate some cases differently. We therefore view AAAP as a scalable, now human-validated route toward automated construction of complex causal-diagnosis benchmarks.
Second, the benchmark scope is limited. CEG-Bench contains 100 traces from three representative agentic benchmarks, covering deep research, function calling, and agentic coding, and all candidate traces are drawn from a single source model (GLM-4.6). We note that the source model only determines which raw traces enter the candidate pool: the difficulty of recovering the gold graph is a property of the trace content, and any error or failure a trace exhibits is expressible within our taxonomy, so changing the generator enlarges the object set rather than altering our conclusions. It does not cover all deployment settings, safety-critical domains, or multi-agent environments. Extending CEG-style diagnosis to more frontier source models and broader tasks remains future work.
Third, repair-value estimation remains preliminary. We emphasize that the \emph{gold} repair values in CEG-Bench are computed by a deterministic graph-structural functional (\S\ref{sec:impl-repair-value}), \emph{not} by an LLM; the calibration difficulty is specific to the inference-time LLM estimate, whose buckets can be affected by model-specific wording and confidence. We therefore recommend the deterministic graph score as the primary prioritization signal and treat the LLM bucket as a fallback; even when bucket thresholds are miscalibrated the underlying ordering can remain useful (CEG-Claude attains Kendall's $\tau=+0.40$, \S\ref{app:exp-repair}). We leave robust cross-backbone repair-value calibration to future work.
Finally, due to computational and API cost constraints, our baseline comparison focuses on Claude Code and Codex, and our cross-backbone study uses five representative LLMs. Broader comparisons across more agent frameworks, LLM families, and deployment environments are left for future work.

\end{document}